\documentclass[conference]{IEEEtran}

\usepackage{amsmath,amssymb,amsfonts}
\usepackage{algorithmic}
\usepackage{graphicx}
\usepackage{textcomp}
\usepackage{xcolor}
\usepackage{bbm}
\usepackage{stmaryrd}
\usepackage{relsize}
\usepackage[hidelinks]{hyperref}
\usepackage[super]{nth}
\usepackage{makecell}
\usepackage[noabbrev]{cleveref}
\usepackage[]{algorithm2e}

\usepackage{xspace}
\newcommand{\hairsp}{\hspace{1pt}}
\newcommand{\ie}{\textit{i.\hairsp{}e.}\xspace}

\newcommand{\wrt}{\textit{w.\hairsp{}r.\hairsp{}t.}\xspace}
\newcommand{\etal}{\textit{et al.}\xspace}

\newcommand{\B}[1]{\mathbf{#1}}

\newcommand{\C}[1]{\mathcal{#1}}
\newcommand{\I}[0]{\mathbbm{1}}

\usepackage{xargs}
\usepackage{mathtools}
\newcommandx\val[3][2,3,usedefault]{{#1_{#2}^{#3}}}

\usepackage{biblatex}
\begin{document}

\title{Rebuilding Trust in Active Learning with Actionable Metrics}


\author{\IEEEauthorblockN{Alexandre \textsc{Abraham}}
\IEEEauthorblockA{\textit{Dataiku} \\
Paris, France \\
alexandre.abraham@dataiku.com}
\and
\IEEEauthorblockN{Léo \textsc{Dreyfus-Schmidt}}
\IEEEauthorblockA{\textit{Dataiku} \\
Paris, France \\
leo.dreyfus-schmidt@dataiku.com}
}

\maketitle

\begin{abstract}
Active Learning (AL) is an active domain of research, but is seldom used in the industry despite the pressing needs. This is in part due to a misalignment of objectives, while research strives at getting the best results on selected datasets, the industry wants guarantees that Active Learning will perform consistently and at least better than random labeling. The very one-off nature of Active Learning makes it crucial to understand how strategy selection can be carried out and what drives poor performance (lack of exploration, selection of samples that are too hard to classify, ...).

To help rebuild trust of industrial practitioners in Active Learning, we present various actionable metrics. 
Through extensive experiments on reference datasets such as CIFAR100, Fashion-MNIST, and 20Newsgroups, we show that those metrics brings interpretability to AL strategies that can be leveraged by the practitioner.
\end{abstract}



\section{Introduction}

Data labeling is a paramount prerequisite to train supervised machine learning algorithms. When labeling is cheap or samples are scarce, all available data can be labeled at once. However, when dealing with larger datasets or when the labeling cost is high, selecting only the most informative samples allows to reach better performance at lesser costs.

This is the promise of Active Learning (AL) which relies on the following iterative process: a model is first trained on already labeled data, then this model and data insights are used to query an oracle with new samples to annotate. This data is added to the labeled pool and the procedure starts again. Real life examples include annotation by domain experts or high-throughput biology \cite{settles2011theories} where acquiring labels takes several hours on a machine that process samples by small batch. The methods proposed to select samples to label are referred to as \textit{informativeness measures} or \textit{query strategies} \cite{settles2009active}. For the sake of simplicity, we will refer to them as strategies in the rest of the paper.

Given that AL experiments can be subject to repetition noise \cite{kottke2017challenges}, a sound practice to compare strategies is to fix all the parameters of the experiment, such as the number of samples to be labelled per iteration. Unfortunately, this experimental design leaves out some crucial practical concerns. For example, the performance is usually evaluated on a left-out labeled test set, that is most often not available in real-life conditions. Even among studies sharing a similar design, it is not rare to observe contradictory results \cite{munjal2020towards}. Finally, not only the AL experiments proposed in the literature differ vastly from real-life scenarios, but the objective of the practitioner is not that of the researcher. If the latter is working for better performance on benchmark datasets, the real-life practitioner is looking for AL methods that are always better than \textbf{Random} labeling.

In this paper, we propose robust settings to run and draw conclusions from active learning experiments together with active learning metrics the ML practitioner can monitor to better assess the quality of the labeling and learning process. We also propose proxies for these metrics, measurable in real-life experiments, and prove that they provide enough insights to help guiding the industrial practitioner.
We present the main concepts of active learning, how they can be quantified, and how those metrics can bring practical information. We also present cardinal, an open-source python package aiming at easing reproducibility in active learning as well as providing better understanding of active learning methods.

\subsection{Active learning strategies}

Settles \cite{settles2009active} presents the foundations of AL techniques and introduces the notion of \textit{sample informativeness}. It is defined as the amount of information a sample brings to a machine learning model when added to its training set. Various methods are proposed to quantify the \textit{informativeness} of a sample: model uncertainty extracts a score from model predictions, expected error reduction measures the impact of samples on the model loss, etc. Du \etal \cite{du2015exploring} refines the notion of sample contribution to a model by distinguishing between \textit{informativeness}, which measures the ability of a sample to reduce the generalization error of a model, and \textit{representativeness} which uses the underlying structure of unlabeled data in order to ensure that the training set is a good representation of the data the model may encounter.

Most of the recent literature on AL aims at refining these concepts and combining them in new ways. \textit{Representativeness} can be defined based on the distance, or similarity, between unlabeled and selected samples \cite{huang2010active, du2015exploring}, as points with smallest inertia \cite{zhdanov2019diverse}, or by counting the inbound edges in a K-nearest neighbor graph \cite{he2017uncertainty}. \textit{Informativeness} and \textit{representativeness} can be optimized together in a single loss \cite{huang2010active, elhamifar2013convex, du2015exploring}, in two-steps by pre-selecting a large batch of samples using uncertainty and then selecting the batch using \textit{representativeness} \cite{zhdanov2019diverse}, or they can be considered as two different arms in a multi-arm bandit setting \cite{bouneffouf2014contextual}.

Finally, pooled active learning, where samples are selected by batches, was developed to avoid the prohibitive cost of retraining the model after each labeling. The notion of batch \textit{diversity} \cite{settles2011theories, du2015exploring} was introduced to limit the redundant information when selecting several samples. It is often considered together with \textit{representativeness} as both are based on sample similarity.

\subsection{Limitations of active learning in practice}

\textbf{Evaluating strategies and reproducing experiments.} 
AL strategies can be compared using Student's t-test on accuracy measures \cite{du2015exploring}, by comparing accuracy at fixed iterations \cite{li2012active}, or simply by visual interpretation of learning curves \cite{zhdanov2019diverse}. However accuracy is very sensitive to the samples selected to start the experiment \cite{kottke2017challenges}, and even on similar tasks with similar designs, some studies contradict each other \cite{munjal2020towards}. This variance together with the wide variety of experimental design in active learning is an impediment to strategy comparison and reproducibility.

\textbf{Incremental vs. fixed test set.} 
Experiments in the AL literature suppose the existence of a large enough and representative testing set for the task at hand, a \textit{fixed test set}. This is an artificial setting that does not represent real-life AL scenarios.
In practice, there is no ground truth labeled set to start with and part of the new labeled data will make the testing set at each labeling iteration. This is the \textit{incremental test set} setting. The model performance evolution can then exhibit non-monotonic behavior in the early stage of the process.

\textbf{Stopping an active learning experiment.} Most studies end after exhausting a pre-defined budget, or until all available samples are labeled, but only few explore the possibility to perform active learning until the accuracy stops improving, or at least not enough to motivate further labeling. Taking this decision requires to know the test accuracy that is not available in a real life experiment. Proxies for this have been proposed such as \textit{contradictions} \cite{he2017uncertainty}, the ratio of samples for which the prediction has changed from one iteration to the other, or the uncertainty score variance \cite{ghayoomi2010using}. We investigate contradictions as a metric of interest in this paper.


\section{Active learning experimental design}

\subsection{Notations and settings}

Let $\mathcal{X}$ be the input space and $\mathcal{Y}$ be the label space where $|\mathcal{Y}|$ is finite for classification tasks. Let $X\in \mathcal{X}$ and $Y\in \mathcal{Y}$ be random variables. Let $d: \mathcal{X} \times \mathcal{X} \xrightarrow{} \mathbbm{R}$ denote any distance over the input space.

Suppose we have an initial pool of $n$ samples composed of a labeled set $\C{D}_{L_{0}} = \{(\val{\B{x}}[1][L_{0}], \val{y}[1][L_{0}]), ..., (\val{\B{x}}[n_{l_0}][L_{0}], \val{y}[n_{l_0}][L_{0}])\}$ independently sampled from $X\times Y$, and an unlabeled set $\C{D}_{U_{0}} = \{\val{\B{x}}[1][U_{0}], ..., \val{\B{x}}[n_{u_0}][U_{0}])\}$ independently sampled from $X$.
At each iteration $t$, samples from $\C{D}_{U_{t-1}}$ are transferred to  $\C{D}_{L_{t}}$ such that $n_{l_{t}}+n_{u_{t}} = n$.
The train dataset is the union $\C{D}_{L_{t}} \cup \C{D}_{U_{t}}$, in which the AL strategy is able to select samples for annotation.

We denote as $\C{D}_{Te} = \{(\val{\B{x}}[1][Te], \val{y}[1][Te]), ..., (\val{\B{x}}[m][Te], \val{y}[m][Te])\}$ the fixed separate test set and remind that its labels $\val{y}[i][Te]$ would not be known in a real-life situation. In a research experimental setting, the ground truth on this left-out set is known and used to measure the accuracy of the classifier. In a real-life setting, we use this left-out set as an unbiased excerpt \wrt query selection. We denote by $h_{t}$ the classifier at iteration $t$ trained on $\C{D}_{L_{t}}$ and class predictions are denoted $h_{t}(\val{\B{x}}[i]) = \val{\hat{y}}[i]$.

\subsection{Active learning metrics}

\textbf{Stopping criterion.} As mentioned in the introduction, we used the \textit{contradiction} metric as a proxy on model improvement. It is the ratio of samples for which the prediction has changed from one iteration to the other.
$$\text{C}(t) = \frac{1}{|\C{D}_{Te}|} \sum_{(\B{x},y) \in \C{D}_{Te}} \I_{[h_{t-1}(\B{x}) \;\neq\; h_{t}(\B{x})]}$$


It can be shown that this ratio is an upper bound of the difference in accuracy between two iterations:
$$\frac{1}{|\C{D}_{Te}|}\left| \sum_{(\B{x},y) \in \C{D}_{Te}} 
\I_{[h_{t-1}(\B{x}) \;=\; y]} - \I_{[h_{t}(\B{x}) \;=\; y]}
) \right|
 \leq \text{C}(t)$$


\textbf{Measuring exploration.} The balance between \textit{representativeness} and \textit{informativeness} mentioned in the introduction may remind the practitioner of the famous exploration-exploitation trade-off. We believe indeed that optimizing for \textit{representativeness} is equivalent to exploring the unlabeled set until the selected samples well captures the distribution of the data. We therefore hypothesize the existence of a minimum set of samples satisfying this condition. AL can then be divided into two regimes: an exploration stage at the beginning and once the data is well explored, an exploitation step to fine tune the classification boundaries. In order to validate this, we need to be able to measure how well the labeled set represents the testing set.

To validate our hypothesis of the two regimes of AL, we need to measure how well the strategy captures the testing set. This is done by the following \textit{Exploration Gradient} score:

$$\text{EG}(t) = \sum_{\B{x} \in \C{D}_{Te}}  d(\val{\B{x}}[][], \C{D}_{L_{t-1}})
-
\sum_{x \in \C{D}_{Te}}  d(\val{\B{x}}[][], \C{D}_{L_{t}})
$$

with $d(\B{x}, \C{D})$ being the distance from $\B{x}$ to its nearest neighbor in $\C{D}$. Note that since we only add labeled sample to $\C{D}$, we have $\C{D}_{L_{t-1}} \subset \C{D}_{L_{t}}$ and therefore $\forall x, d(\val{\B{x}}[][], \C{D}_{L_{t-1}}) \geq d(\val{\B{x}}[][], \C{D}_{L_{t}})$  which means that this metric is strictly decreasing.


\textbf{Easy-to-classify samples.} Related to the concept of \textit{informativeness}, we introduce the notion of difficulty to classify samples. This notion has been known for a long time, in particular in the context of semi-supervised learning where samples that are easy to classify are assigned a (pseudo) label but samples hard to classify are not automatically labeled \cite{dasgupta2009mine}. To the best of our knowledge, this is the first time that this criterion is used in an active learning context to explain the performance of query strategies. We hypothesize that easy-to-classify samples bring little to no information to the model while hard-to-classify samples, if selected too often, may be detrimental to the classifier. To identify such hard-to-classify samples, we propose to train the same model on the test set. Samples where the classifier is mistaken are probable hard-to-classify samples. This is measured on a batch by:
$$\text{reverse batch accuracy}(\C{D}_B) = \frac{1}{|\C{D}_B|} \sum_{(\B{x},y) \in \C{D}_B} \mathbbm{1}_{h_{\C{D}_Te}(\B{x}) = y}$$

In real-life settings where the test set is unlabeled, we propose as a proxy the agreement between the classifier $h$, and a 1-nearest-neighbor classifier $h_{N\!N}$. This is denoted by $\kappa$ and is computed after each iteration on $\C{D}_B$:

$$\kappa(\C{D}_B) = \frac{1}{|\C{D}_B|} \sum_{x \in \C{D}_B} \mathbbm{1}_{h(x) = h_{N\!N}(x)}$$

\section{Experiments}

In order to evaluate active learning strategies, we have selected various tasks where the label could actually be retrieved by a human, such as image labeling or determining if a website is malicious.
We use tabular datasets from openML that represent easy tasks, \ie reaching very high accuracy with only few samples, similar to the ones used in other papers \cite{du2015exploring, huang2010active} to validate the interest of active learning and purpose of the proposed metrics. We have also selected widely used bigger datasets, such as \textbf{CIFAR-100} and \textbf{LDPA}, for which it is more tricky to determine whether one strategy dominates, and if we can infer it based only on metrics.

\subsection{Protocol}

Our study compares the performance of several AL strategies on a panel of datasets and run an aggregated statistical comparison of those methods to determine whether some strategies are significantly better than others. We use area under the accuracy curve as performance metric similarly to \cite{cardoso2017ranked}. All of our figures displays the average over all folds along with 10th and 90th quantile of values.

The number of repeated experiments varies across AL studies and can go from one to a hundred \cite{kottke2017challenges}. For our study, we follow the recommendation of \cite{kottke2017challenges} and repeat our experiments ten times. Dietterich \cite{dietterich1998approximate} argues that 5 replications of 2-fold cross-validation is the most powerful repetition scheme with acceptable Type I error. This approach in interesting as it results in having training and testing sets of the same size.

In order to compare results across tasks and strategies, we cannot assume that results are drawn from a normal distribution, and therefore ANOVA cannot be used.
Following \cite{demvsar2006statistical}, we use a Friedman test to determine the best methods over all datasets. Since we compare each method against all the others, we use a Nemenyi post-hoc test when the null hypothesis is rejected to ensure that the difference is significant. In the figures, CD denotes the critical distance necessary for two variant to be statistically significantly different ($\text{p} \leq 0.05$). All studies and figures were performed using the \textrm{autorank} package \cite{herbold2020autorank}.

We also want to measure the correlation between our actionable metrics, which are reliable and measured using ground truth, and their real-life proxy. We use Pearson's correlation coefficient when the metrics values are meaningful. When the values are less relevant and we only want to compare the ranking, we use a Spearman's rank correlation coefficient.

\subsection{Datasets description and processing}

We divide our pool of datasets into two categories. The so-called easy tasks comes from tabular datasets where a good performance is reached within few iterations, without any tuning and are used as sanity checks. Those tasks are completed by harder tasks to measure generalizability of the proposed approach. All tabular datasets have been preprocessed using one hot encoding for categorical variables and standardization for numerical values. \Cref{tab:datasets} provides a description of all datasets and their associated experimental setting.

For the easy tasks\footnote{~Not to be mistaken with easy samples evoked before.}, we use a RandomForest with scikit-learn's default parameters (version 0.23). \textbf{NOMAO} \cite{candillier2012design} is a location deduplication dataset. The annotator must determine if two locations returned by the search engine are similar or not. It is slightly imbalanced and has both numerical and categorical features.
Phishing Websites \cite{mohammad2012assessment}, abbreviated \textbf{Phishing}, is a website classification task based on network and search-engine-based features. It has two balanced classes and consists in categorical features.
Wall-following robot navigation \cite{freire2009short}, abbreviated \textbf{Robot}, looks for the optimal action of a robot to explore a room. It has 4 classes and, in a real life setting, we could imagine data to be labeled using a simulator.

Harder tasks require more expertise, however, for the sake of simplicity, we have decided to use off-the-shelf methods with default hyperparameters.
\textbf{LDPA} is a tabular dataset composed of values gathered by accelerometers worn by a person performing 11 activities. The goal is to classify the activity based on the sensor values. For this problem we use a Random Forest similar to easy tasks.
20 newsgroups dataset, abbreviated \textbf{20 NG}, is a text classification dataset where messages coming from an online newsgroup must be classified according to their category. We vectorize the text using a pipeline of scikit-learn's CountVectorizer and TfIdfTransformer and then learn a scikit-learn SGDClassifier.
\textbf{MNIST} and Fashion-MNIST, abbreviated \textbf{Fashion}, are well known image classification datasets. They are both composed of 28x28 grayscale images. \textbf{MNIST} represents handwritten digits and is therefore easier to classify compared to \textbf{Fashion} that represents clothes. For both tasks we using a multi-layer perceptron with hidden layers of size 128 and 64.  
\textbf{CIFAR-10} and \textbf{CIFAR-100} \cite{krizhevsky2009learning} are image classification datasets. We compute embeddings for both datasets by taking the output of the last convolutional layers aggregated with an average pooling.
For both classification tasks, we use a multi-layer perceptron with layers 128 and 64.

\begin{table}
  \caption{Summary table of the datasets used in our experiments and corresponding active learning settings. The first group refers to easy tasks, and the second one to hard tasks.}
  \label{tab:datasets}
  {\centering
  \begin{tabular}{l|c|c|c|c|c}
    Name & Classes & Samples & Start size & Batch size & Steps \\
    \hline
    \textbf{NOMAO} & 2 & 34 465 & 10 & 20 & 20 \\
    \textbf{Phishing} & 2 & 11 055 & 20 & 50 & 20 \\
    \textbf{Robot} & 4 & 5 456 & 10 & 15 & 15 \\
    \hline
    \textbf{MNIST} & 10 & 60 000 & 100 & 100 & 10 \\
    \textbf{Fashion} & 10 & 60 000 & 100 & 100 & 10 \\
    \textbf{CIFAR-10} & 10 & 60 000 & 1000 & 1000 & 10 \\
    \textbf{CIFAR-100} & 100 & 60 000 & 1000 & 1000 & 10 \\
    \textbf{LDPA} & 11 & 164 000 & 100 & 100 & 30 \\
    \textbf{20 NG} & 20 & 20 000 & 100 & 100 & 20 \\
  \end{tabular}}
\end{table}

\subsection{Query strategies}

Our baseline strategy for query selection is the \textbf{Random} strategy. We also use the classic uncertainty strategies based on the model prediction probabilities as defined in \cite{settles2009active}:
\begin{equation}
\begin{split}
\textbf{Confidence}(\B{x})& = 1 - \B{\hat{y}}^1\\
\textbf{Margin}(\B{x})& = 1 - (\B{\hat{y}}^1 - \B{\hat{y}}^2)\\
\textbf{Entropy}(\B{x})& = - \sum_i \log(\B{\hat{y}}^i)\B{\hat{y}}^i\\
\end{split}
\end{equation}
with $\B{\hat{y}}^i$ the $i$\textsuperscript{th} predicted probability sorted in descending order.

Taking inspiration from \cite{zhdanov2019diverse}, we use a \textbf{KMeans} based strategy that performs a clustering with \textit{batch size} clusters and selects the samples closest to the cluster centroids. This unsupervised method aims at maximizing exploration.
Finally, \textbf{WKMeans} is the strategy referred to as WClustered(10) in \cite{zhdanov2019diverse} that expresses a trade-off between uncertainty and exploration. This method uses the \textbf{Margin} uncertainty strategy to first select 10 times the desired \textit{batch size}, and then applies the \textbf{KMeans} strategy weighted by the \textbf{Margin} scores.

\subsection{Performance on easy and hard tasks}

\begin{figure}
    \centering
    \includegraphics[width=\linewidth]{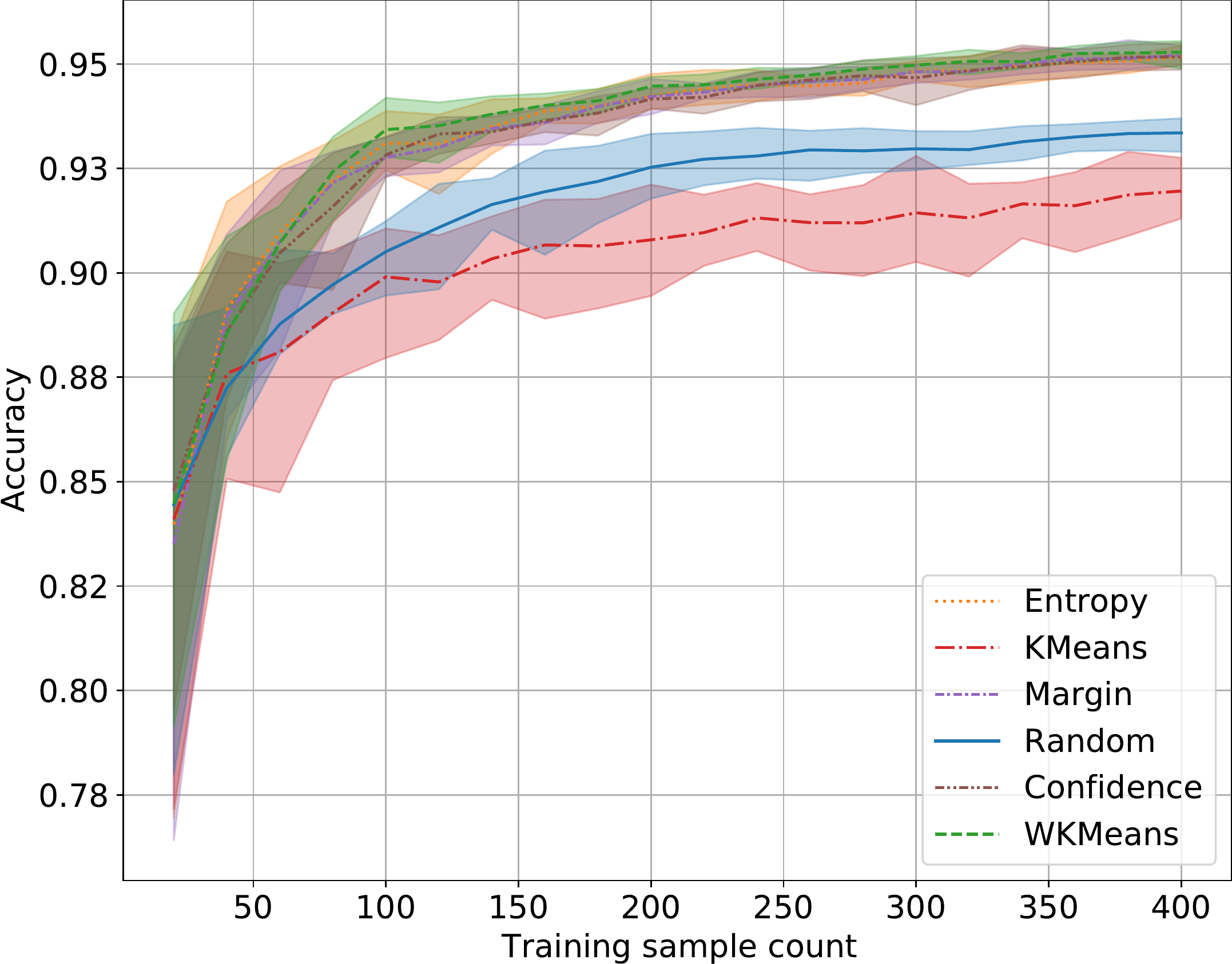}
    \caption{Comparison of strategies on \textbf{NOMAO} dataset}
    \label{fig:nomao_accuracy}
\end{figure}

\begin{figure}
    \centering
    \includegraphics[width=\linewidth]{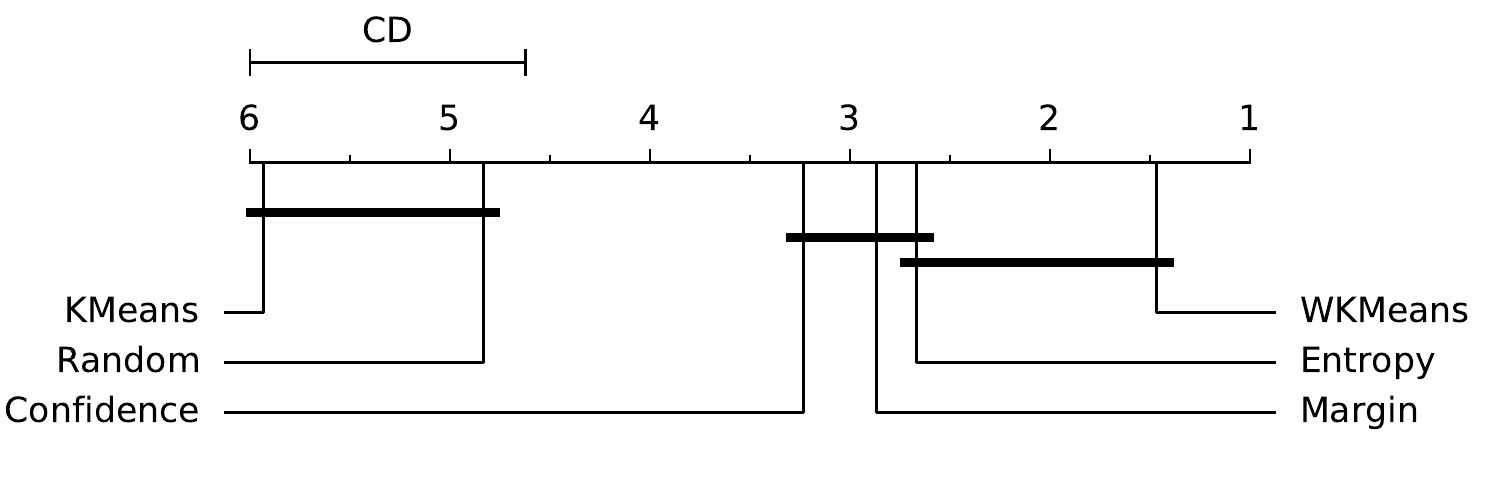}
    \caption{Comparison of methods on easy tasks}
    \label{fig:method_ranking_small}
\end{figure}

\Cref{fig:nomao_accuracy} shows the accuracy of AL strategies on the \textbf{NOMAO} dataset. We chose this task as it has the smallest repetition noise. Results for the other datasets are available in the \cref{app:nomao_accuracy,app:phishing_accuracy,app:robot_accuracy}. \textbf{Random} and \textbf{KMeans} are clear underperforming strategies on this dataset, but the repetition noise prevents from selecting a best one.
\Cref{fig:method_ranking_small} presents an analysis of strategy ranks on easy tasks. From our results, we conclude that supervised strategies perform well enough on easy tasks to justify a deeper analysis of the results. We would advise the practitioner to pick any uncertainty strategy for that kind of task. 
Those easy tasks not being representative of real-life scenarios, we now move on to the harder tasks to see whether those preliminary results generalize.

\begin{figure}
    \centering
    \includegraphics[width=\linewidth]{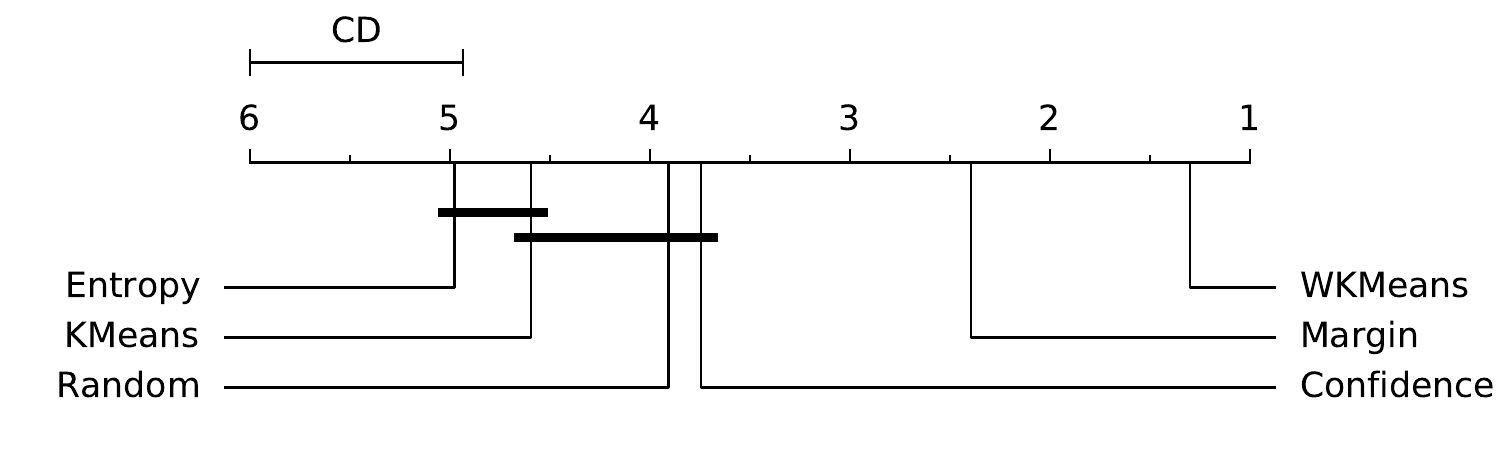}
    \caption{Comparison of methods on hard tasks}
    \label{fig:method_ranking_big}
\end{figure}

Harder tasks exhibit more complicated patterns, as seen in \cref{fig:ldpa_accuracy_contradiction}. For an in-depth analysis of the particularity of each task, we refer the reader to detailed results on all dataset in \cref{app:cifar10_accuracy,app:cifar100_accuracy,app:mnist_accuracy,app:fashion_accuracy,app:ldpa_accuracy,app:news_accuracy}.
\Cref{fig:method_ranking_big} shows the strategy comparison on all hard tasks. Compared to easy tasks, \textbf{Random} and \textbf{KMeans} are performing better, while the rank of \textbf{Entropy} and \textbf{Confidence} has dropped, which highlights that exploration may be more important on harder tasks.
\textbf{Margin} has improved a little and dominates all other methods just below \textbf{WKmeans}. This result was surprising for us as \textbf{Margin} is not the default uncertainty-based method in many papers \cite{elhamifar2013convex, cardoso2017ranked}. 

\subsection{Contradictions as stopping criterion}

\begin{figure*}
    \centering
    \includegraphics[width=.48\linewidth]{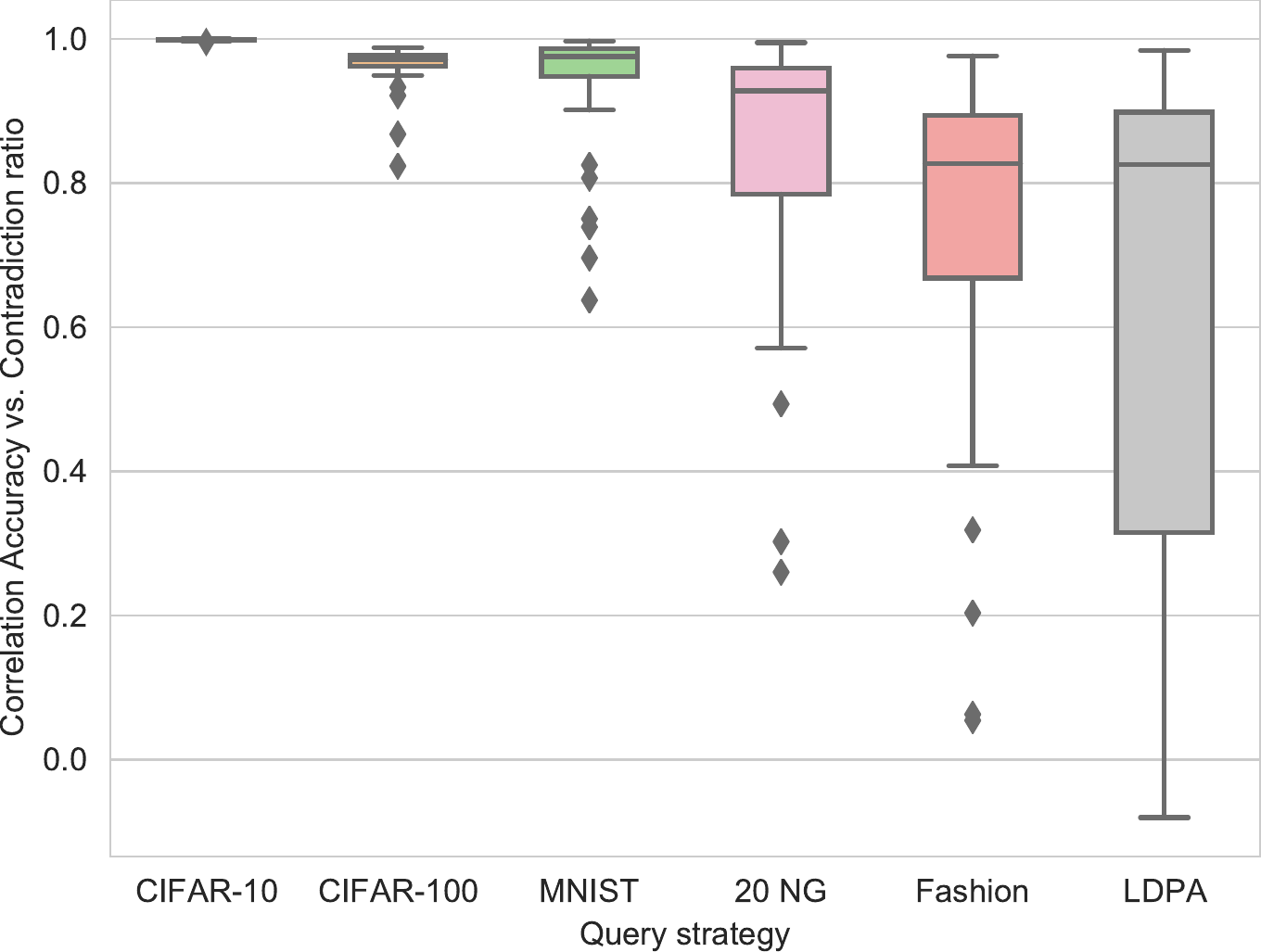}
    \includegraphics[width=.48\linewidth]{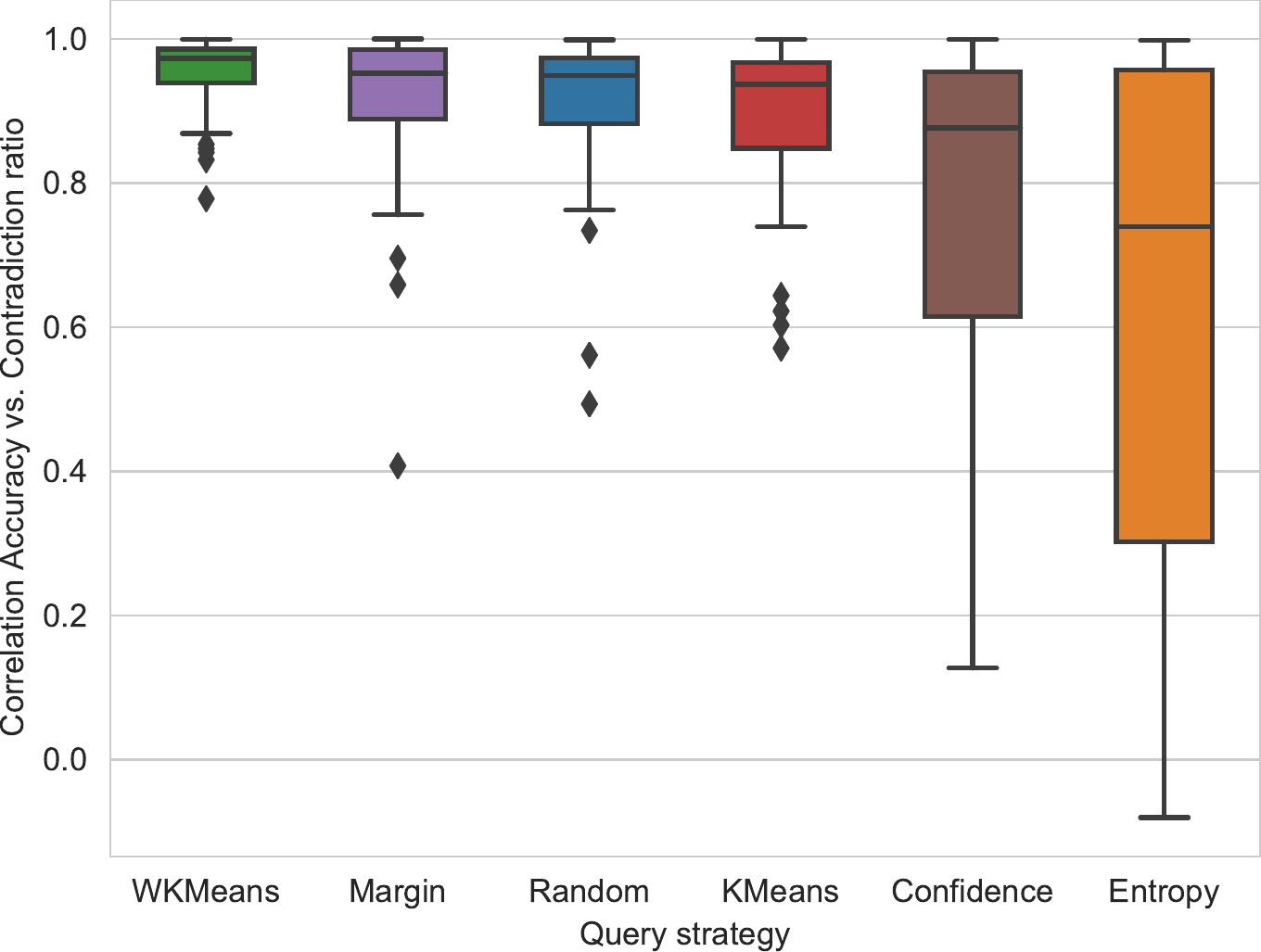}
    \caption{Correlations between accuracy and contradictions. \textit{Left.} By dataset. \textit{Right.} By method.}
    \label{fig:corr_accuracy_contradiction}
\end{figure*}

\begin{figure*}
    \centering
    \includegraphics[width=.48\linewidth]{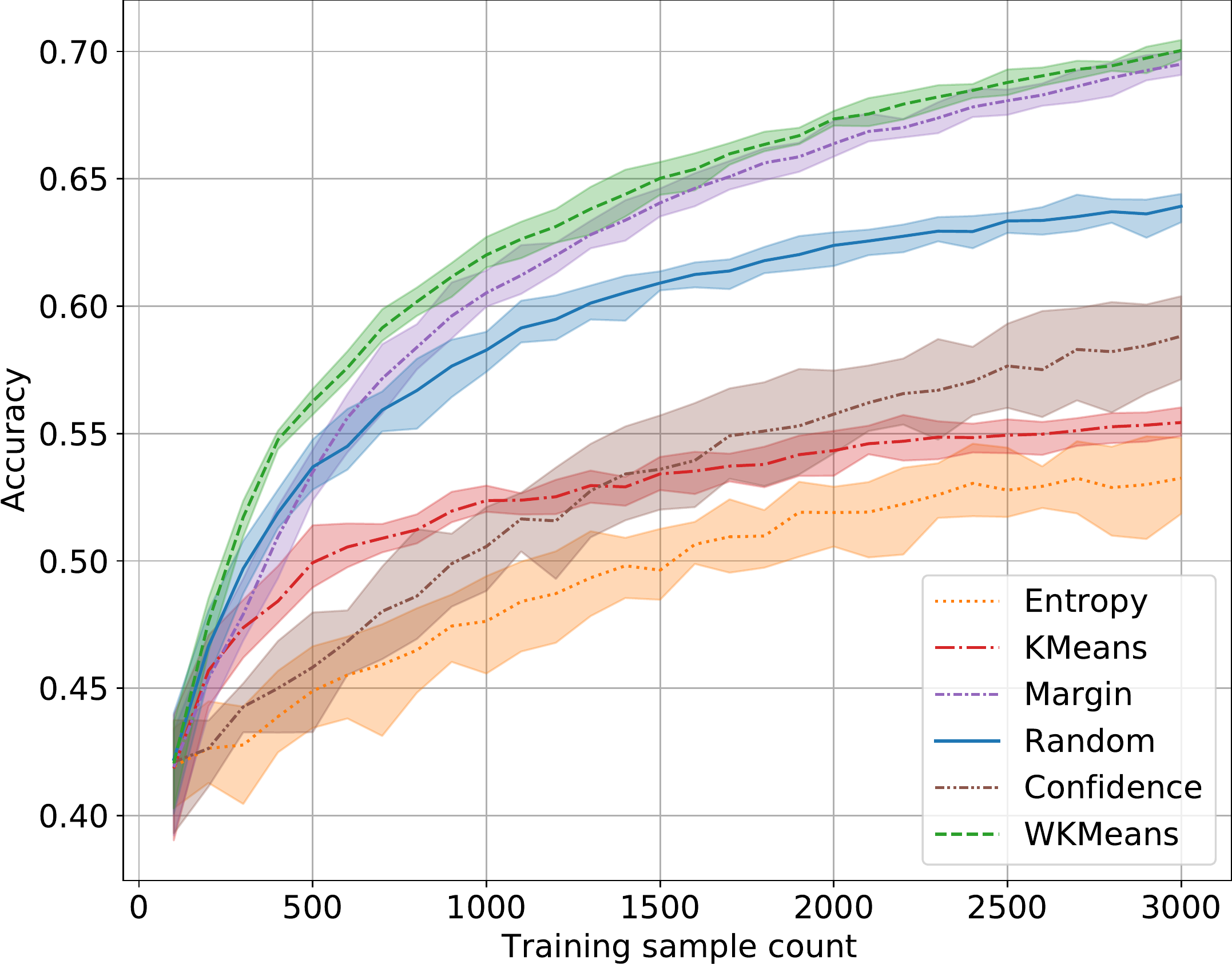}
    \includegraphics[width=.48\linewidth]{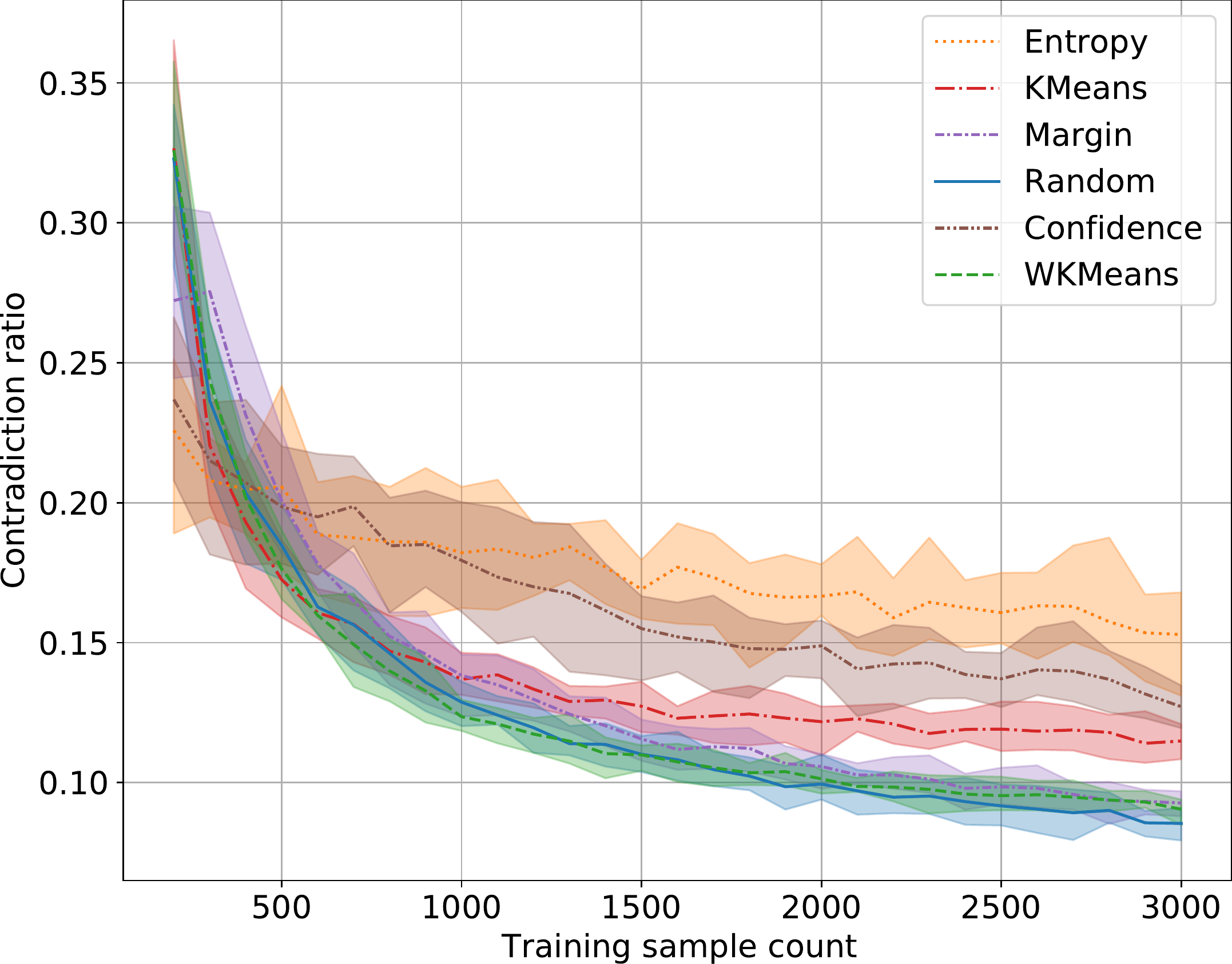}
    \caption{\textit{Left.} Accuracy of different strategies on LDPA dataset. \textit{Right.} Contradictions measured on the same experiment.}
    \label{fig:ldpa_accuracy_contradiction}
\end{figure*}

We noted that, by construction, the ratio of contradictions measured from one iteration to the next on the test set is an upper bound of the accuracy gradient. However, this measure can be inefficient if the new prediction is not accurate, or worse if the classifier was right originally but changed.
\Cref{fig:corr_accuracy_contradiction} displays the correlation between accuracy and contradictions by strategy and by dataset. We  notice a high correlation on datasets \textbf{CIFAR-10}, \textbf{CIFAR-100}, and \textbf{MNIST}, and on all methods but \textbf{Entropy} and \textbf{Confidence}. This result is encouraging since \textbf{WKmeans} and \textbf{Margin} are the best performing strategies and both show high correlations. \Cref{fig:ldpa_accuracy_contradiction} focuses on \textbf{LDPA} where the results are the less convincing. We observe on these figures that \textbf{Entropy} and \textbf{Confidence} are performing badly while having a high number of contradictions: this indicates that the classifier changes its prediction for an incorrect label. 

\subsection{Measuring exploration}

\begin{figure}
    \centering
    \includegraphics[width=\linewidth]{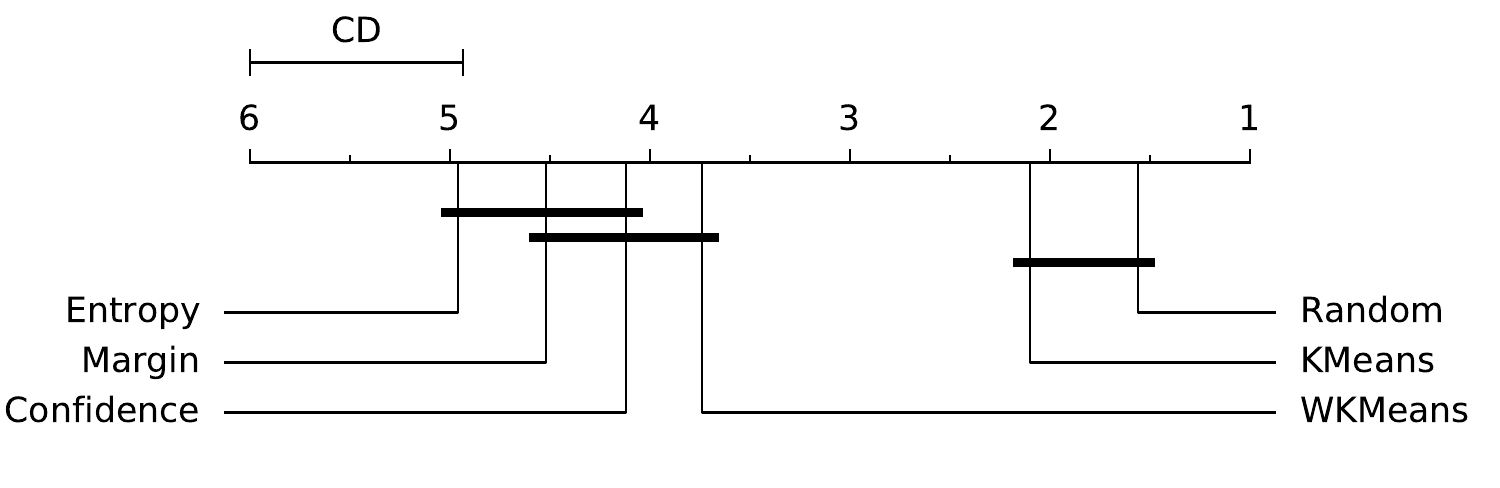}
    \caption{Comparison of AUC of exploration scores.}
    \label{fig:exploration_ranking}
\end{figure}

\begin{figure}
    \centering
    \includegraphics[width=\linewidth]{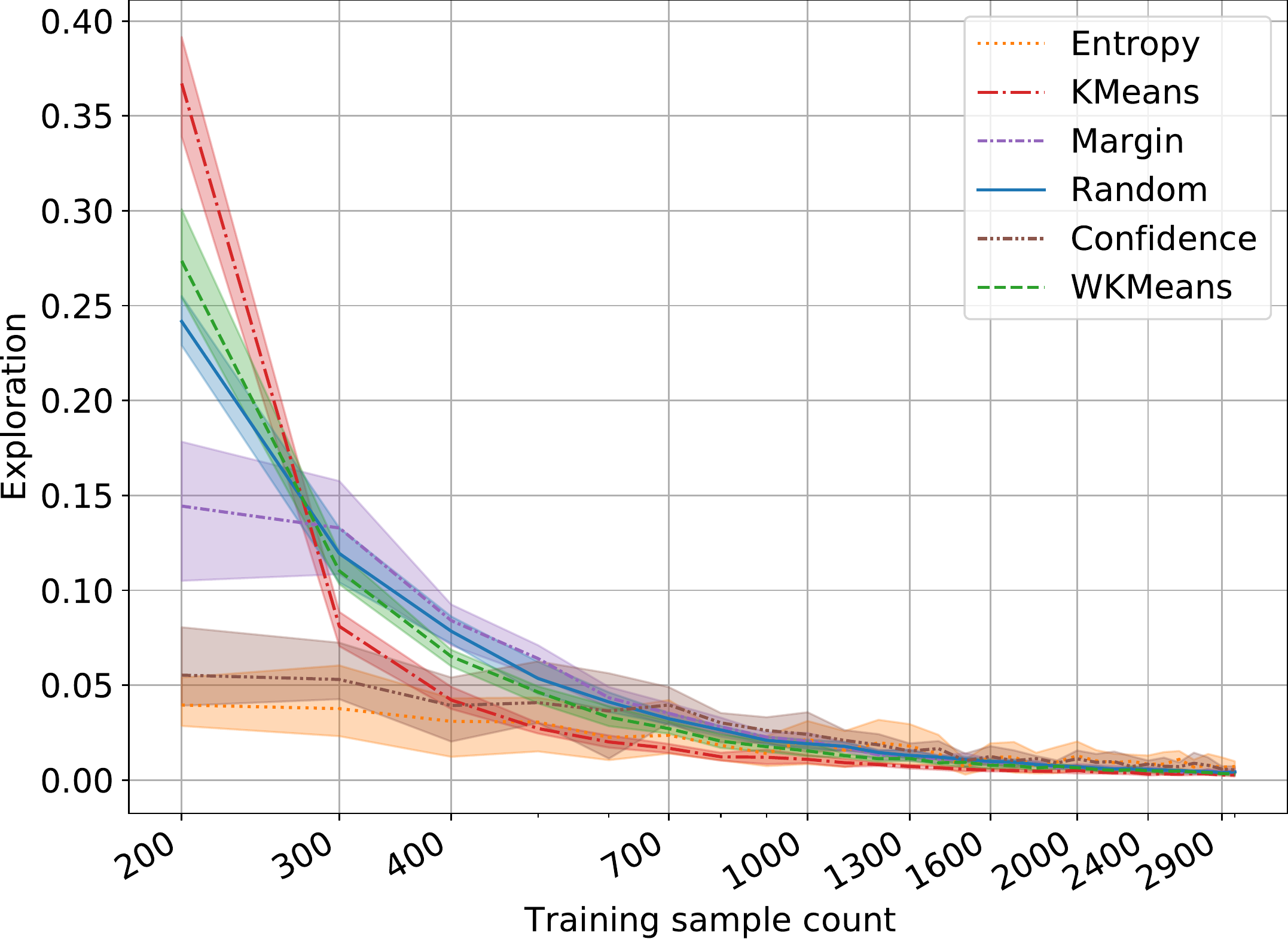}
    \caption{Exploration metric on \textbf{LDPA}.}
    \label{fig:ldpa_top_exploration}
\end{figure}

\Cref{fig:ldpa_top_exploration} shows that the greatest differences in exploration metric are only visible for the first iterations, we therefore highlight them using a log scale. We confirm that \textbf{Entropy} and \textbf{Confidence} are the least exploring techniques. Since it is hard to conclude from visual inspection only, we run a statistical test across all hard tasks comparing the AUC of the exploration metrics (the individual curves for all tasks are available in \cref{app:cifar10_exploration,app:cifar100_exploration,app:mnist_exploration,app:fashion_exploration,app:ldpa_exploration,app:news_exploration,app:nomao_exploration,app:phishing_exploration,app:robot_exploration}).
\Cref{fig:exploration_ranking} shows the ranking of the methods. As expected, \textbf{KMeans} and \textbf{Random} explore way more than other methods since they are not encouraged to select samples on the decision boundary of the classifier. Even though it is not significant, we note that \textbf{WKMeans} is the one that weights exploration the most among other methods.

\subsection{$\kappa$ as reverse batch accuracy estimator}
\label{ssec:agreement}
Despite very high accuracy rates, classifiers trained on a large amount of data can still make mistakes. The samples on which they make wrong predictions are typically samples on which model uncertainty is high. We expect uncertainty-based strategies to select more hard samples that methods optimizing for diversity, and observe if this is a useful insights for active learning.
We compute this metric in our experiments and check that $\kappa$ is a good proxy for it.

\begin{figure*}
    \centering
    \includegraphics[width=.48\linewidth]{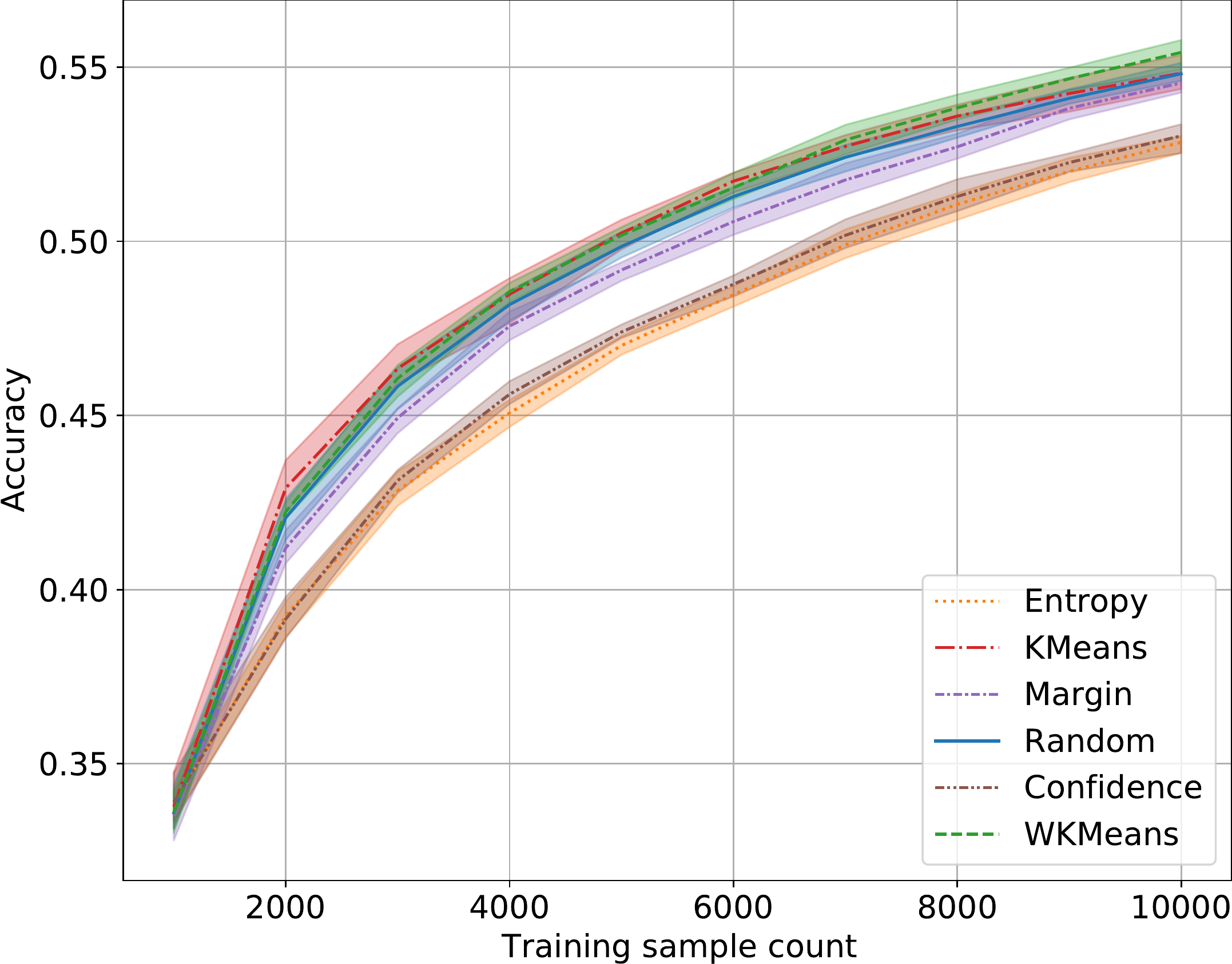}
    \includegraphics[width=.48\linewidth]{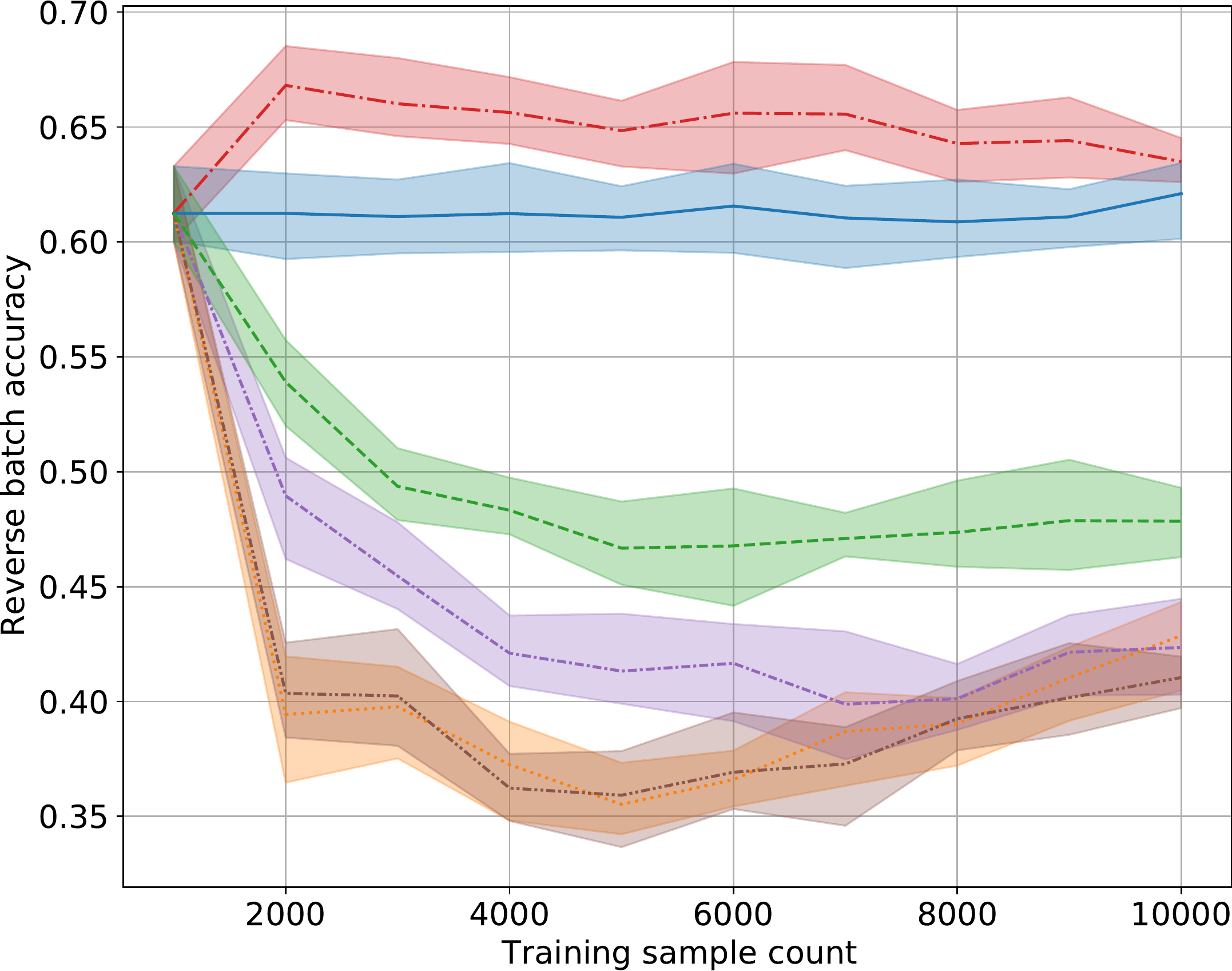}
    \caption{Global accuracy of different strategies (\textit{left}) and reverse batch accuracy (\textit{right}) on \textbf{CIFAR-100}. Legends are the same for both plots.}
    \label{fig:cifar100_accuracy_difficulty}
\end{figure*}

\begin{figure*}
    \centering
    \includegraphics[width=.48\linewidth]{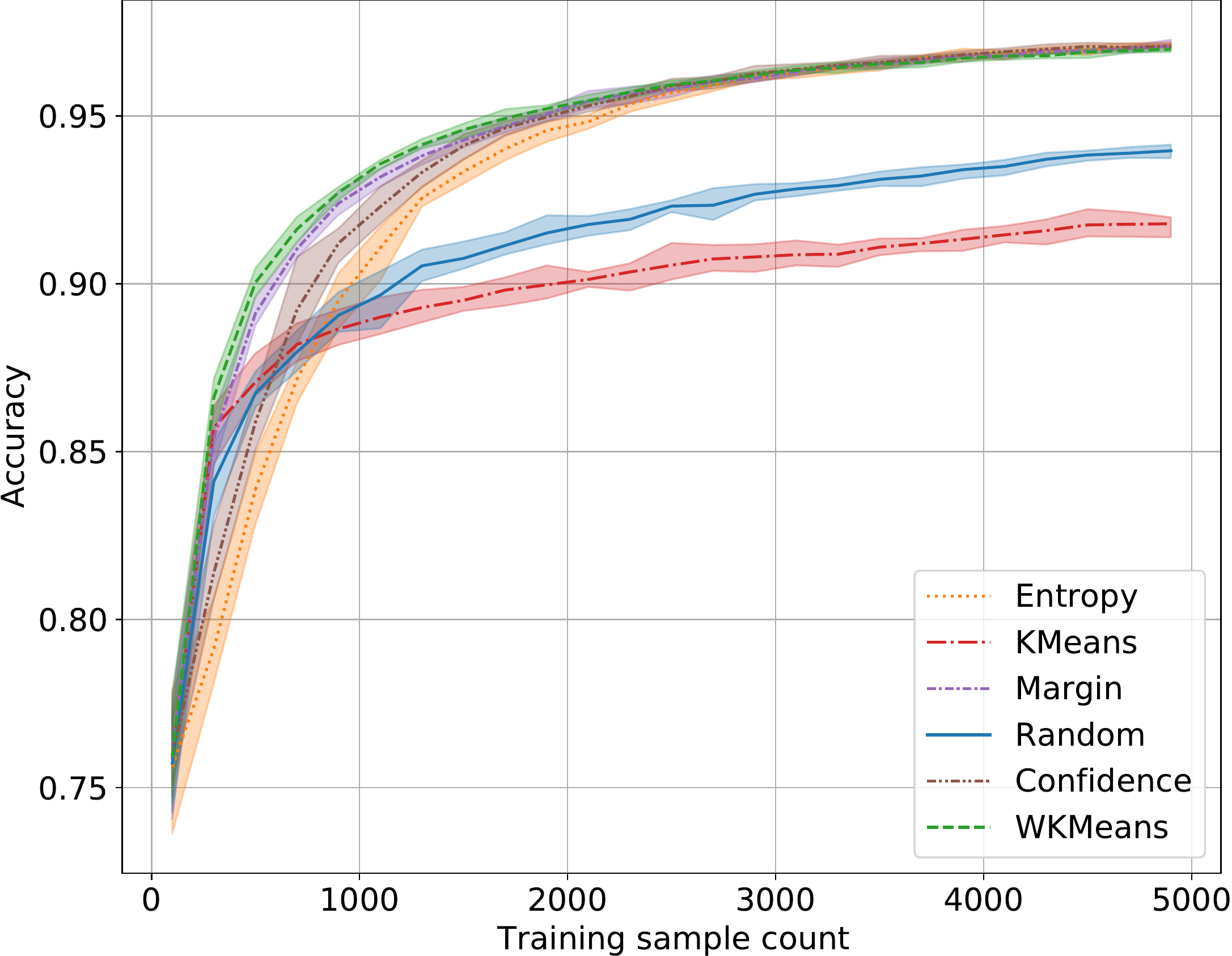}
    \includegraphics[width=.48\linewidth]{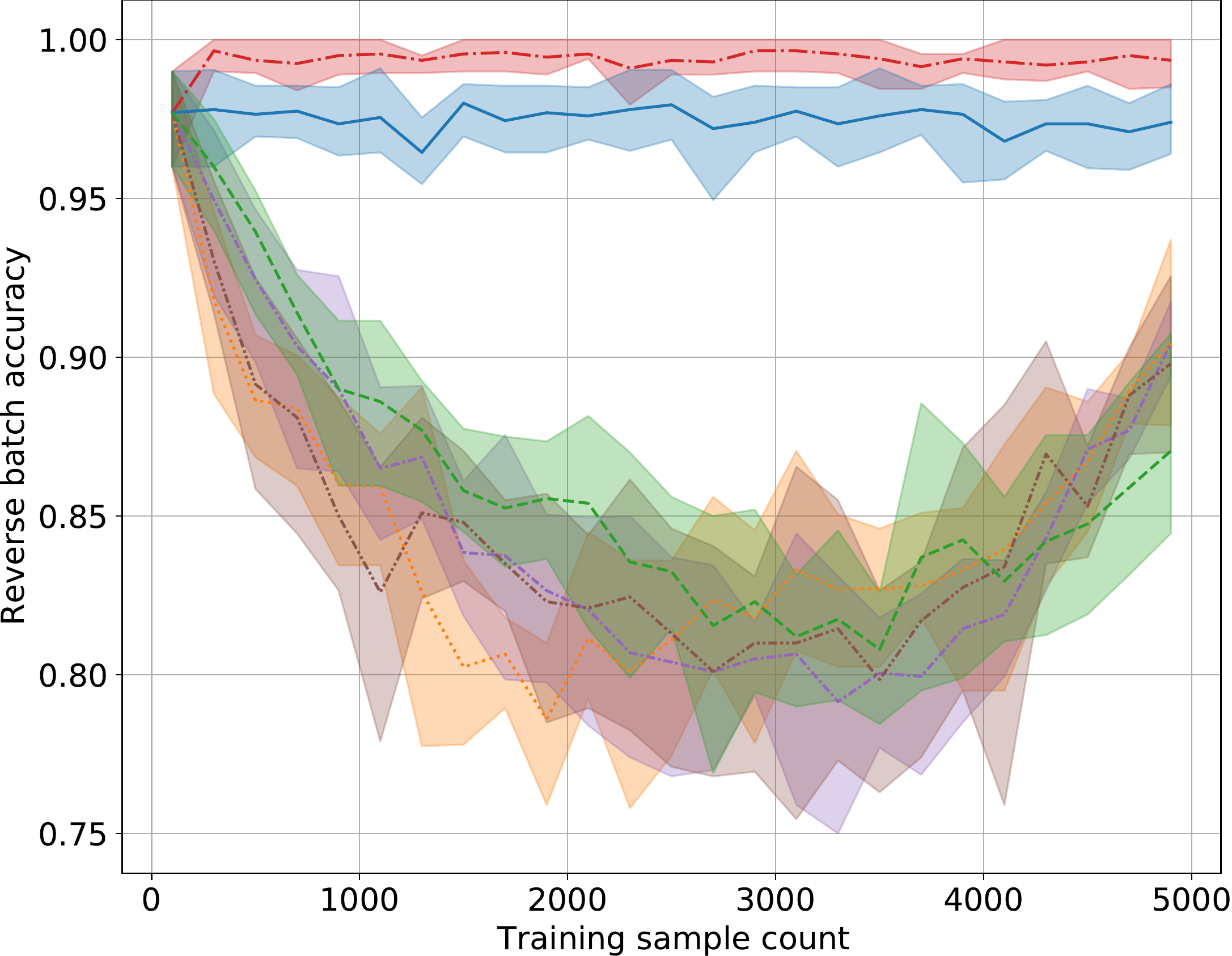}
    \caption{Global accuracy of different strategies (\textit{left}) and reverse batch accuracy (\textit{right}) on \textbf{MNIST}. Legends are the same for both plots.}
    \label{fig:mnist_accuracy_difficulty}
\end{figure*}

\begin{figure}
    \centering
    \includegraphics[width=\linewidth]{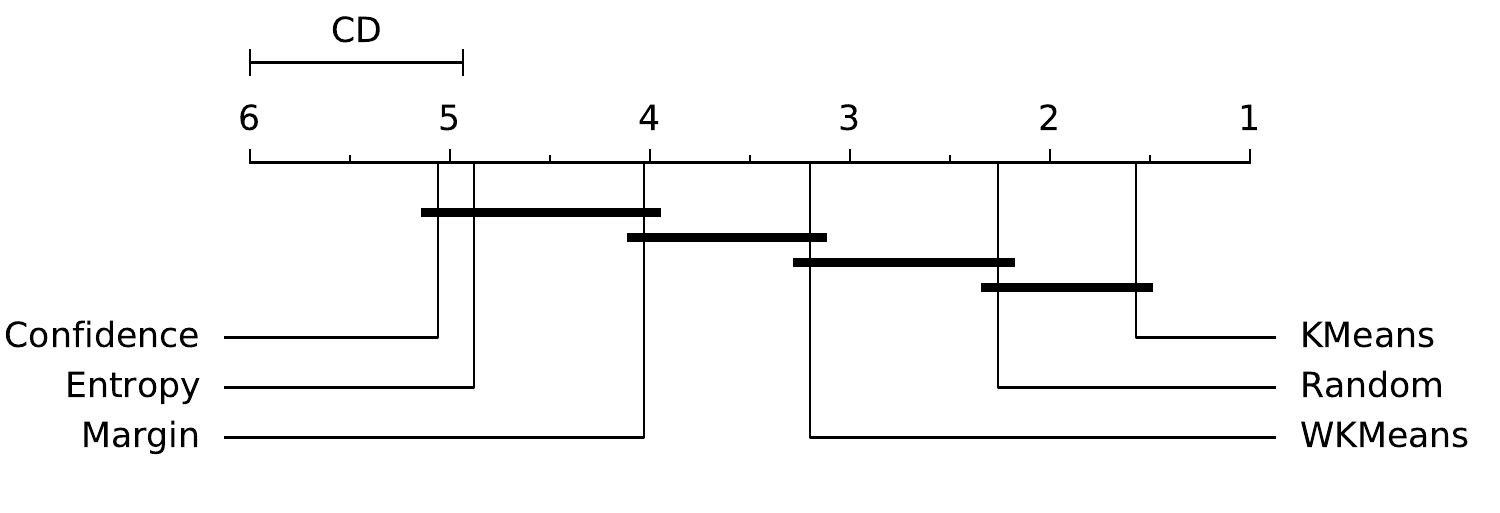}
    \caption{Comparison of the AUC of reverse batch accuracy across strategies.}
    \label{fig:difficulty_ranking}
\end{figure}

\Cref{fig:cifar100_accuracy_difficulty} displays accuracy on \textbf{CIFAR-100} along with reverse batch accuracy. In this experiment, \textbf{Entropy} and \textbf{Uncertainty} strategies select the highest ratio of hard samples among all methods, which could explain their bad performance.
\Cref{fig:mnist_accuracy_difficulty} shows the accuracy on \textbf{MNIST} and reveals an unexpected behavior. The underperforming strategies, \textbf{KMeans} and \textbf{Random}, are selecting predominantly easy samples. Their poor performance could be explained by the fact that easy samples may not provide enough \textit{informativeness} as they are very likely already correctly classified.
To confirm these observations, we have run a comparison of AUC of reverse batch accuracy across all hard tasks.
This resulting CD diagram in \cref{fig:difficulty_ranking} shows that unsupervised methods select significantly easier samples than uncertainty-based ones. \textbf{WKMeans}, which is in-between, has no significant difference with the other methods. In the end, it appears that constantly selecting easy samples or too hard samples is not ideal. There exists a sweet spot where \textbf{WKMeans}, which has the best performance, lies. Based on this observation, when considering a panel of several strategies, we recommend to avoid the ones that select always very easy or very hard samples.

\begin{figure}
    \centering
    \includegraphics[width=\linewidth]{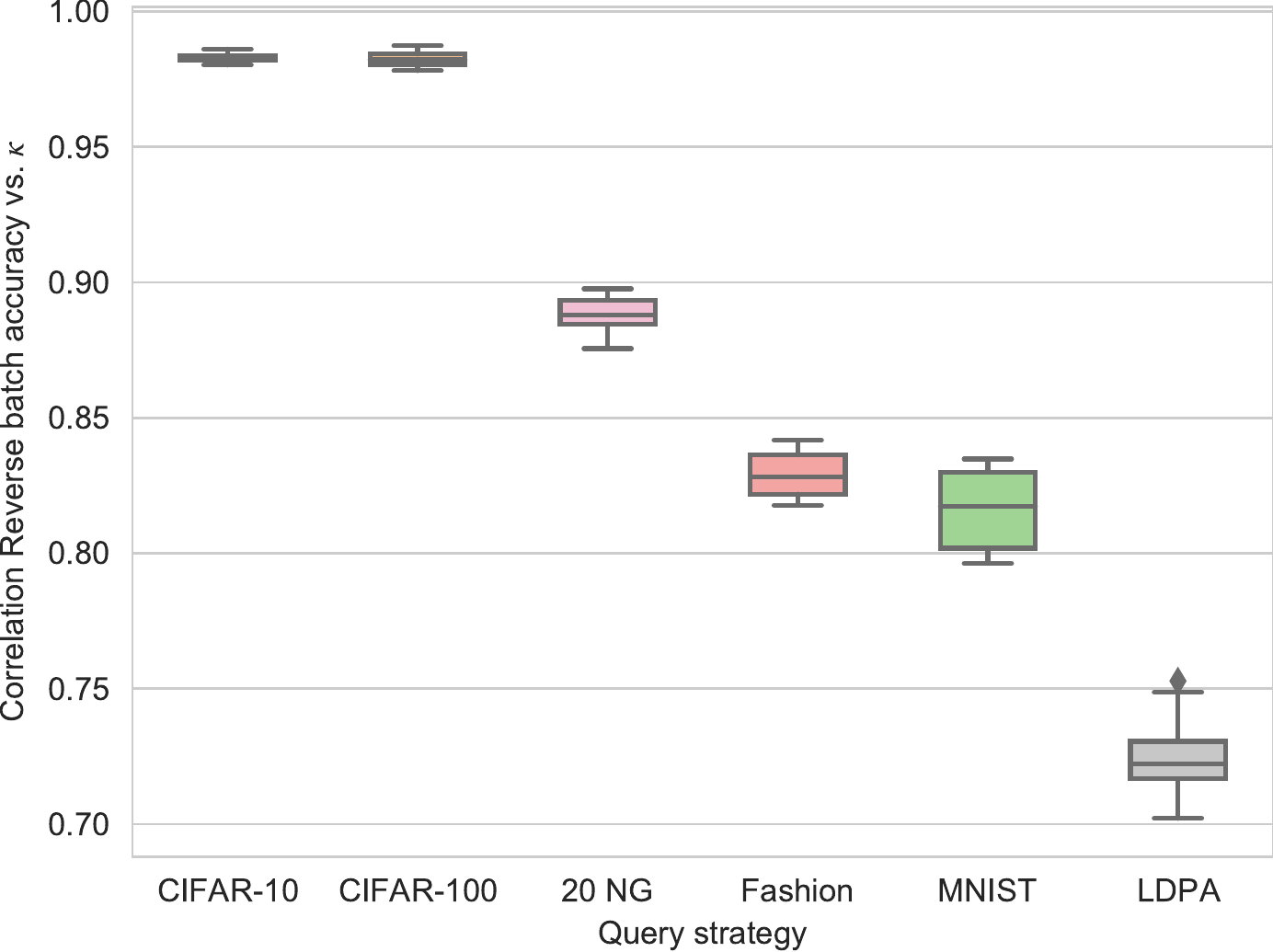}
    \caption{Correlation between the reverse batch accuracy and $\kappa$ across datasets.}
    \label{fig:corr_agreement_difficulty}
\end{figure}

Reverse batch accuracy is hard to assess and, in a real-life setting without any point of comparison, it is hard to interpret. Therefore, we would like for this metric to determine the rank of the methods rather
than their absolute values. To determine this, we measure the rank matching between $\kappa$ and reverse batch accuracy using Spearman's rank correlation coefficient. \Cref{fig:corr_agreement_difficulty} shows a high correlation between metrics for most datasets, with the lowest correlation being LDPA.

\section{A new framework to ease reproduciblity and metric computation}

Existing active learning frameworks focus on providing a wide variety of strategies. Our approach is different as we aim at distributing tried-and-tested strategies, insightful metrics, and reproducible experiments. For this, we have decided to create our own open-source python package called cardinal.

\subsection{Experimental approaches in existing AL frameworks}

modAL \cite{danka2018modal} is a lightweight package that features the most common strategies through Python functions along with a wrapper object that exposes them as scikit-learn estimators. Following its minimalistic philosophy, modAL aims at using the simplest approach sometimes at the cost of scaling. It provides a very simple object destined to handle the for-loop of the active learning experiment but it does not go beyond that.
Libact \cite{yang2017libact} is a performance oriented package. It exposes a Python interface but relies on high speed C code for some of its method. Although it does not provide a full active learning experiment framework, it provides helpers to make the construction of the loop easier.
ALiPy \cite{tang2019alipy} is a package focused on experimentation. It provides an experimenter object that allows to run classic AL experiment with fixed batch size and bounded cost. Its snapshot ability allows to resume a stopped experiment but not to compute additional metrics afterward. It provides classical machine metrics and an analyzer to plot them and perform t-test. However, all these methods and metrics are provided with no guidance which makes it hard to decide which one to use.

\subsection{Cardinal: Trustworthy Active Learning in Python}

The goal of cardinal\footnote{~\url{http://dataiku-research.github.io/cardinal}} is to help perform robust and reproducible AL experiments  \cite{munjal2020towards} by leveraging various simple query strategies and actionable metrics. At the moment, cardinal is still at an early stage. It features all the query strategies mentioned in this study.

\textbf{Evaluating AL strategies.} Designing and testing Active Learning metrics is a continuous work in progress. We have explored several metrics but decided to only release the ones reported in this study that have been proven particularly useful in practice. We also make available our evaluation framework that relies on statistical tests and correlation measures following machine learning best practices \cite{demvsar2006statistical, dietterich1998approximate}.


\textbf{Promoting research and reproducibility.} AL experiments are composed of a succession of not-so-costly steps but that can generate a lot of intermediate data such as learned models, predicted probabilities for all samples, etc. Our framework persists short term cache needed to resume an interrupted AL experiment, along with long term cache that allows post-hoc metric computation. After an experiment has run, cardinal allows to restore the state of the experiment at each iteration to compute a new metric. On the technical side, we log all metrics values in a SQLite database to allow fast and concurrent access, and store large variables on disk by file. Cardinal promotes reproducibility by offering the scripts to reproduce Zhdanov's study \cite{zhdanov2019diverse} along with this study.  Unfortunately, our experiment framework  is  not  mature  enough  to  be  released and is isolated on its own branch for now.



\section{Conclusion}

Active Learning is a complex experimental design that leaves no room for error since experiments are one shot. Its applicability in real life has been questioned in particular given the high variability of results reported for similar experiments \cite{munjal2020towards} or even across repetitions \cite{kottke2017challenges}.

We propose to use actionable metrics during experiments allowing to gain insight on AL strategies and better explain the performance gaps. We also introduce proxies on these metrics that can be monitored in real life experiments by the practitioner and prove their usefulness. For example, batch agreement can be used to spot when a strategy is performing badly because it is selecting samples too hard to classify.
Based on these metrics, we proposed an evaluation framework to rank strategies and guide the practitioner. It is made available in cardinal, our practical AL python package with a strong focus on experiment, along with all the scripts necessary to reproduce this study. 

Further research will focus on using the actionable metrics to select the best strategy during an AL experiment, and in particular how to combine exploration and exploitation metrics. We aim at providing an AL online framework able to change the strategy during an experiment to improve accuracy.

\printbibliography
\clearpage
\appendix
\vspace{-2em}

\setcounter{figure}{0}
\renewcommand{\thefigure}{\thesection.\arabic{figure}}

\newcommand{\accuracy}[2]{
    \begin{figure*}[htbp]
    \centering
    \includegraphics[width=.48\linewidth]{figures/#1_accuracy.pdf}\hfill
    \includegraphics[width=.48\linewidth]{figures/#1_hard_contradiction.pdf}
    \caption{Accuracy and contradiction ratio for \textbf{#2}}
    \label{app:#1_accuracy}
    \end{figure*}
}

\accuracy{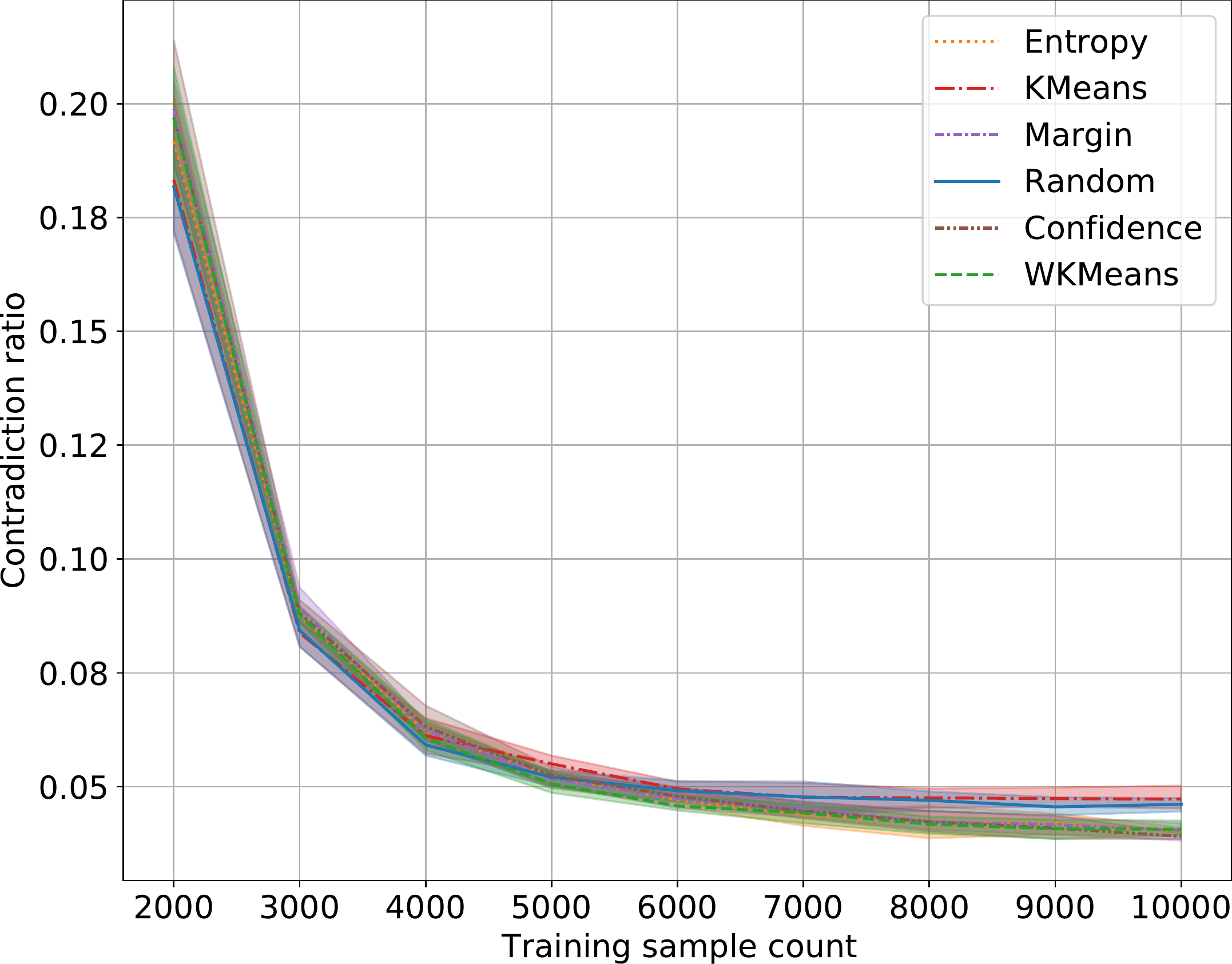}{CIFAR-10}
\accuracy{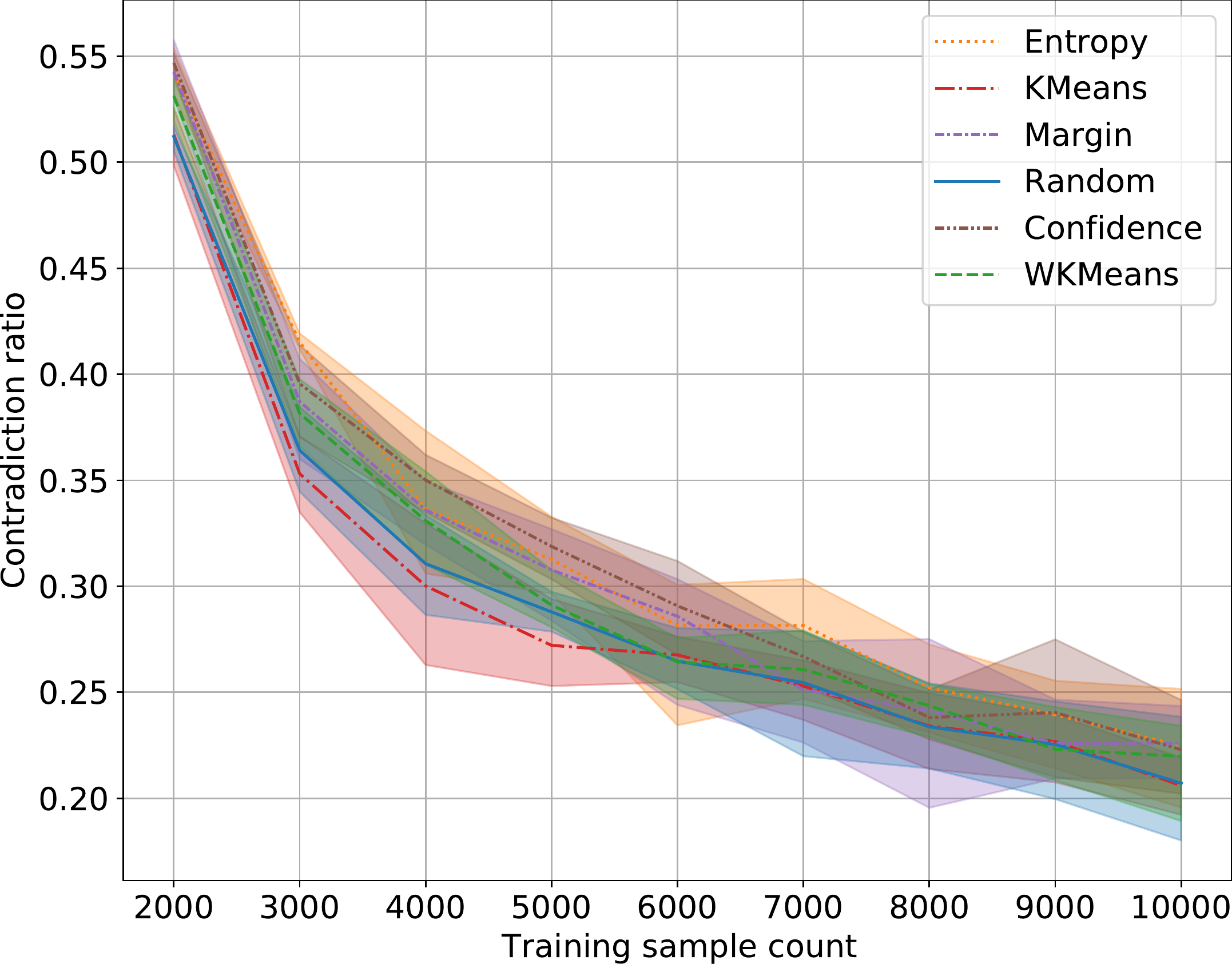}{CIFAR-100}
\accuracy{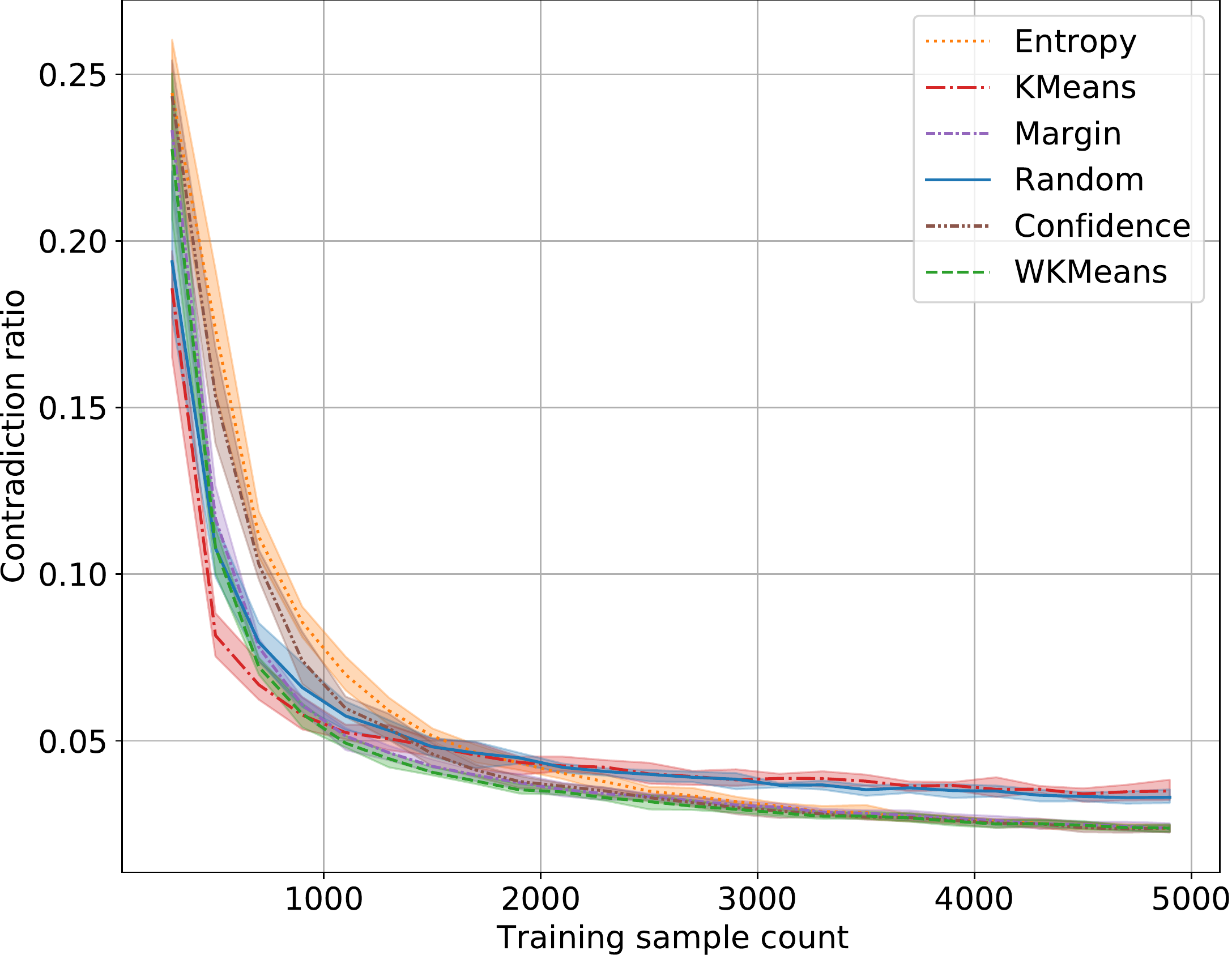}{MNIST}
\accuracy{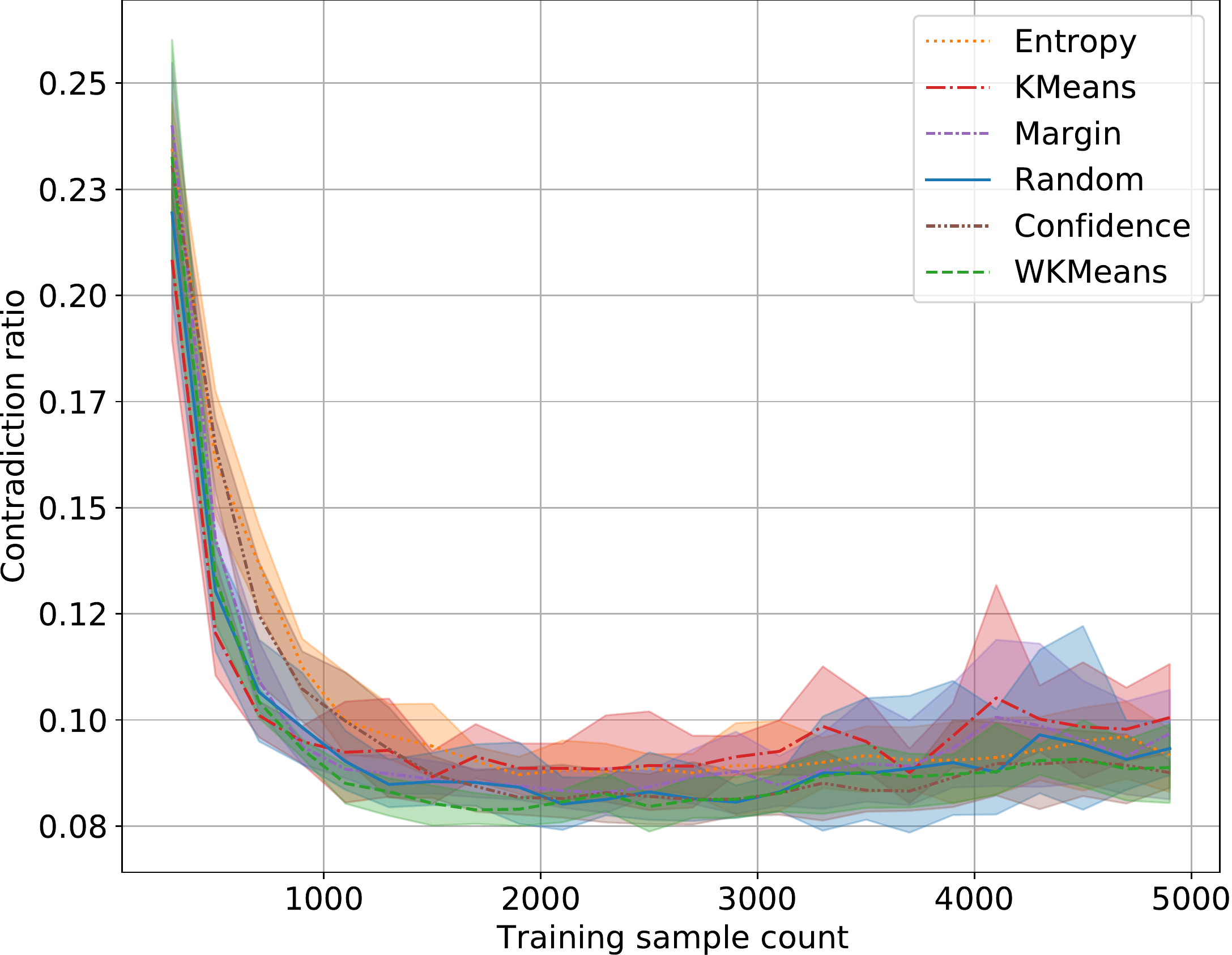}{Fashion}
\accuracy{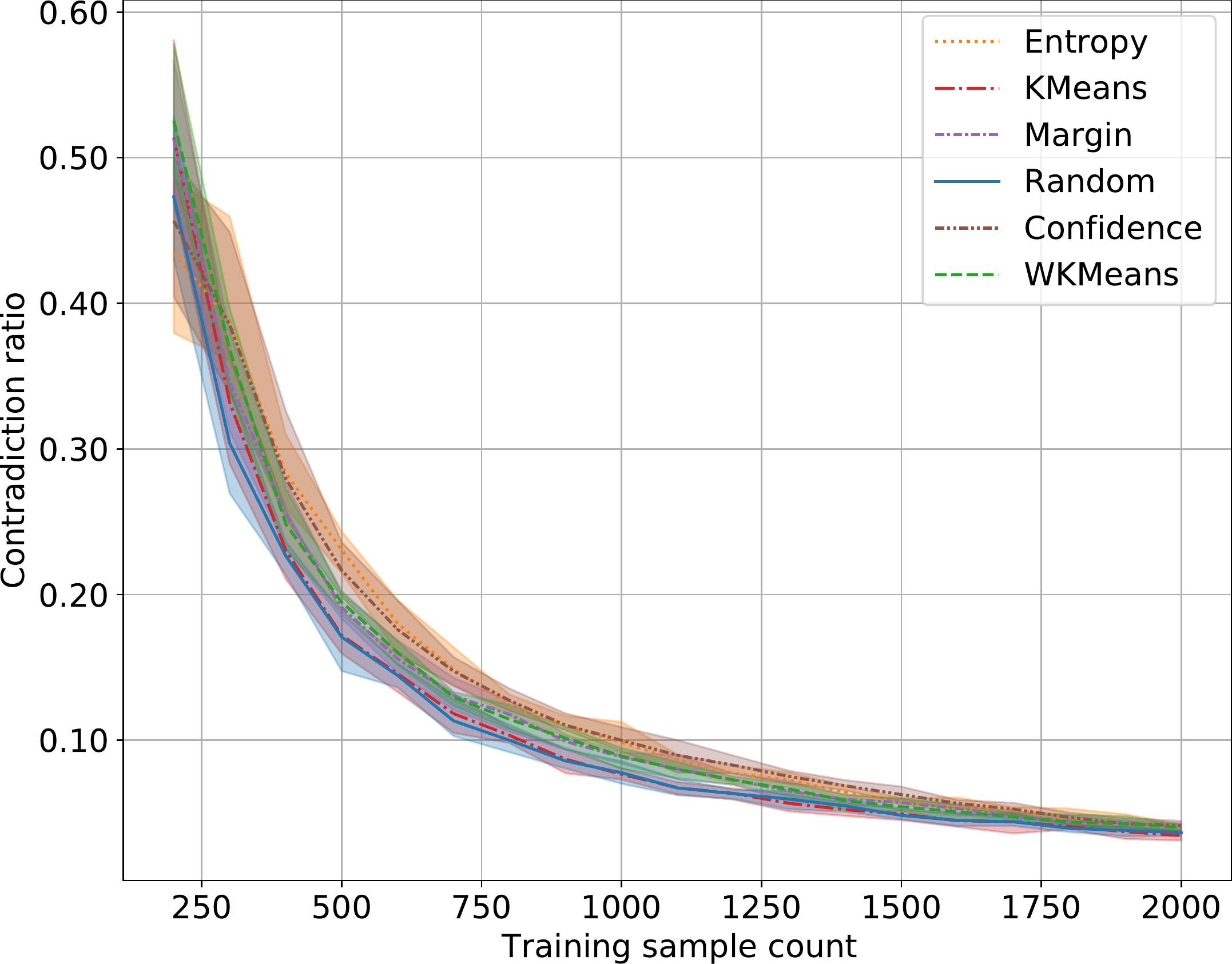}{20 NG}
\accuracy{ldpa}{LDPA}
\accuracy{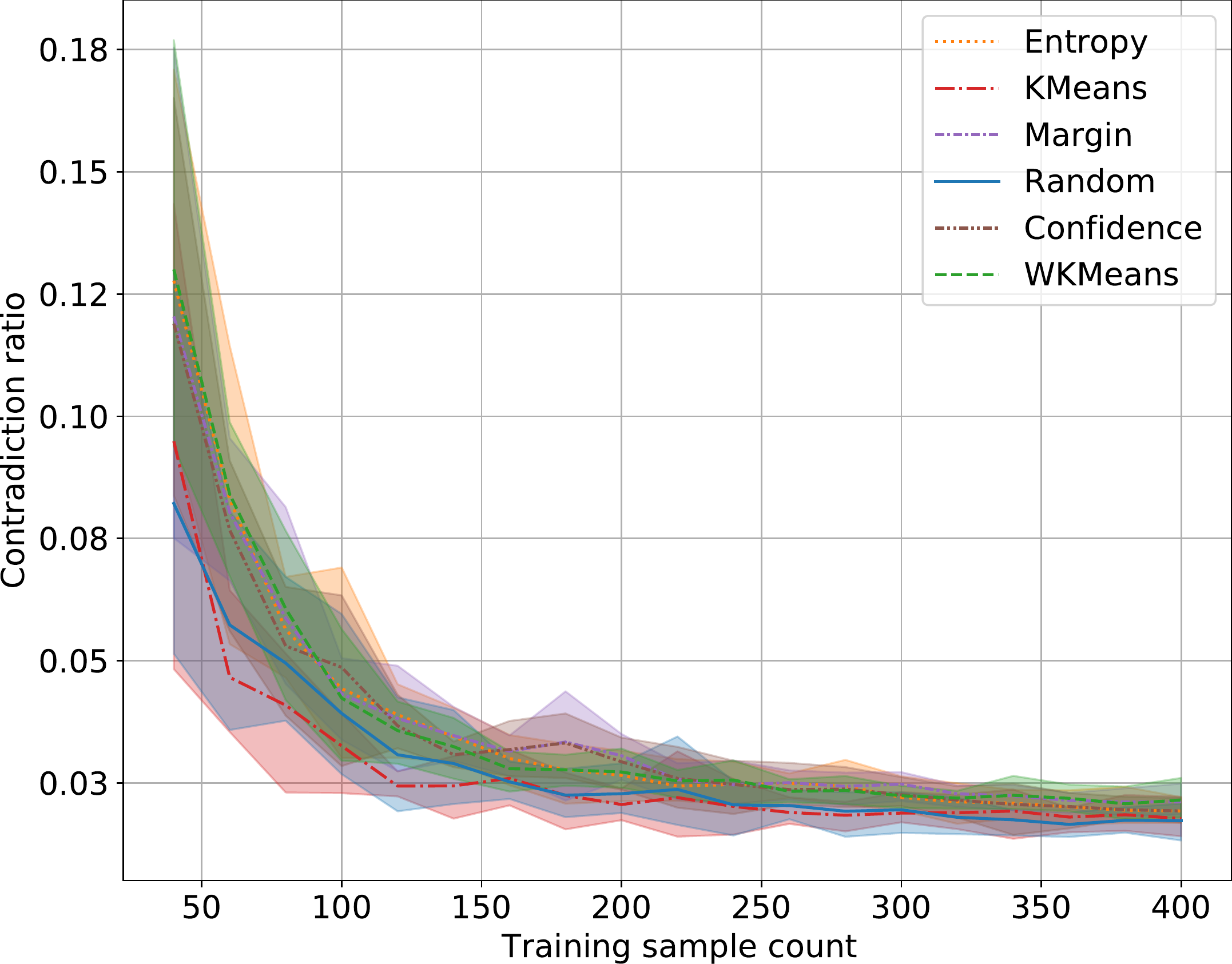}{NOMAO}
\accuracy{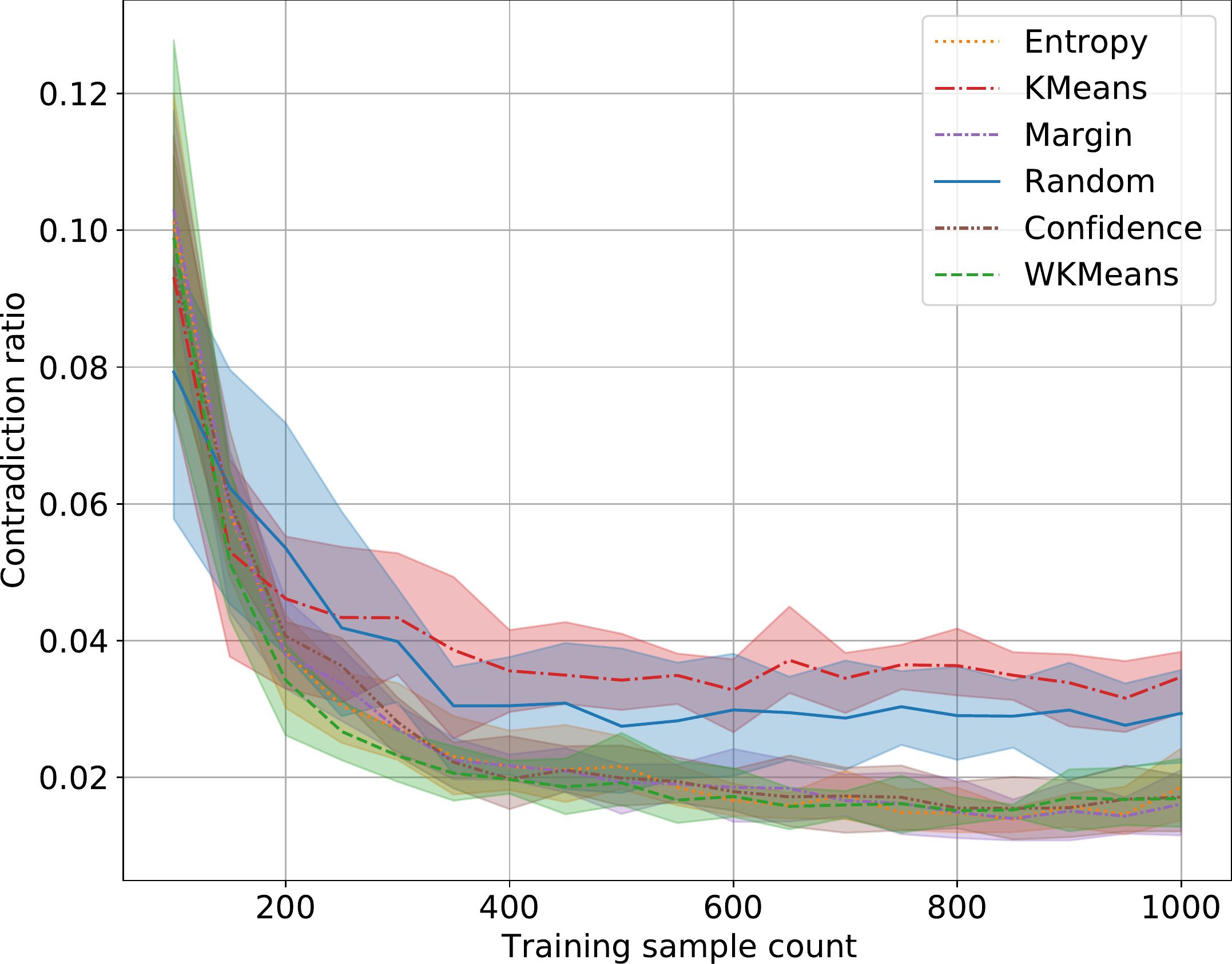}{Phishing}
\accuracy{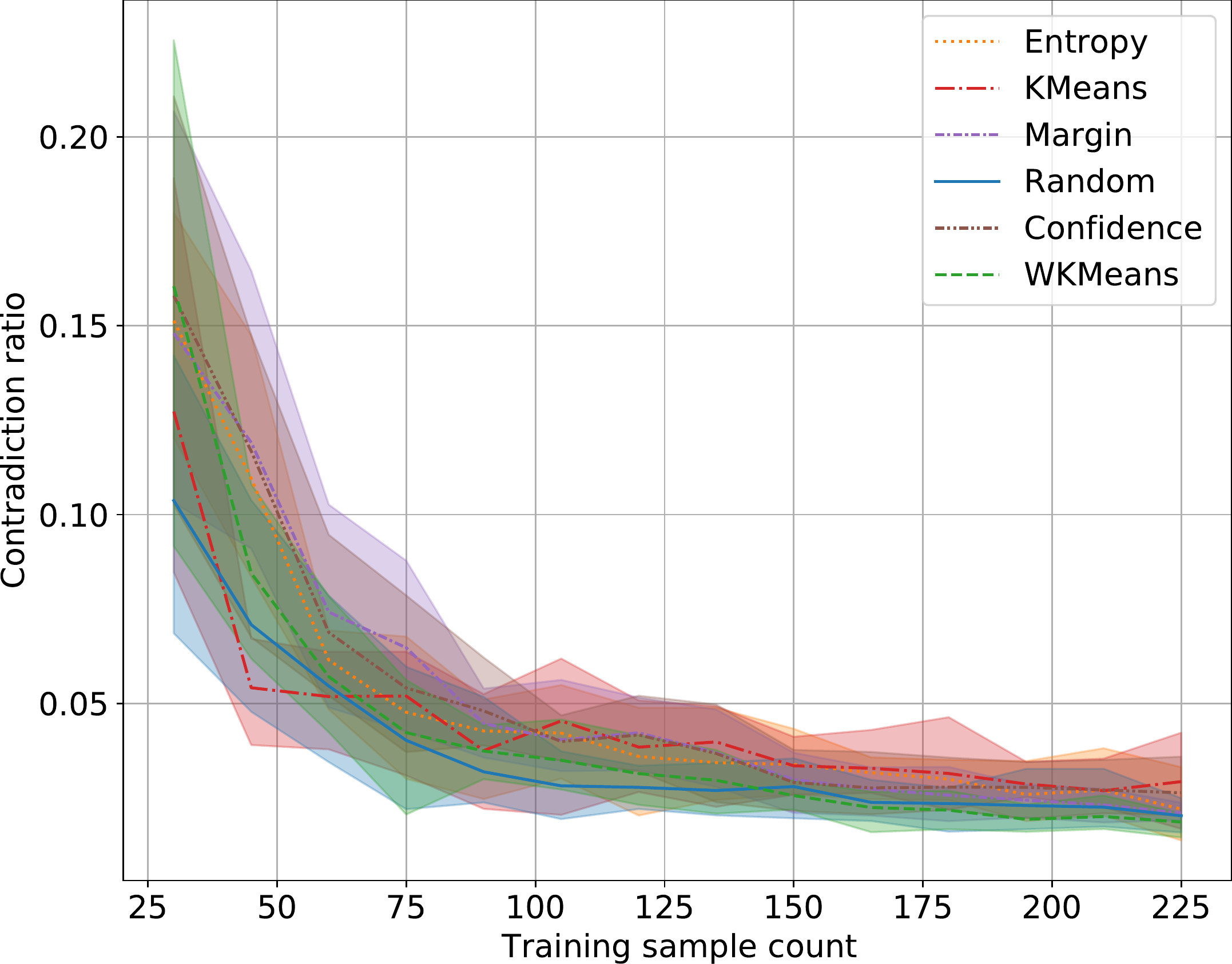}{Robot}

\newcommand{\exploration}[2]{
    \begin{figure}
        \includegraphics[width=\linewidth]{figures/#1_top_exploration.pdf}
        \caption{Exploration score for the \textbf{#2} dataset}
        \label{app:#1_exploration}
    \end{figure}
}

\exploration{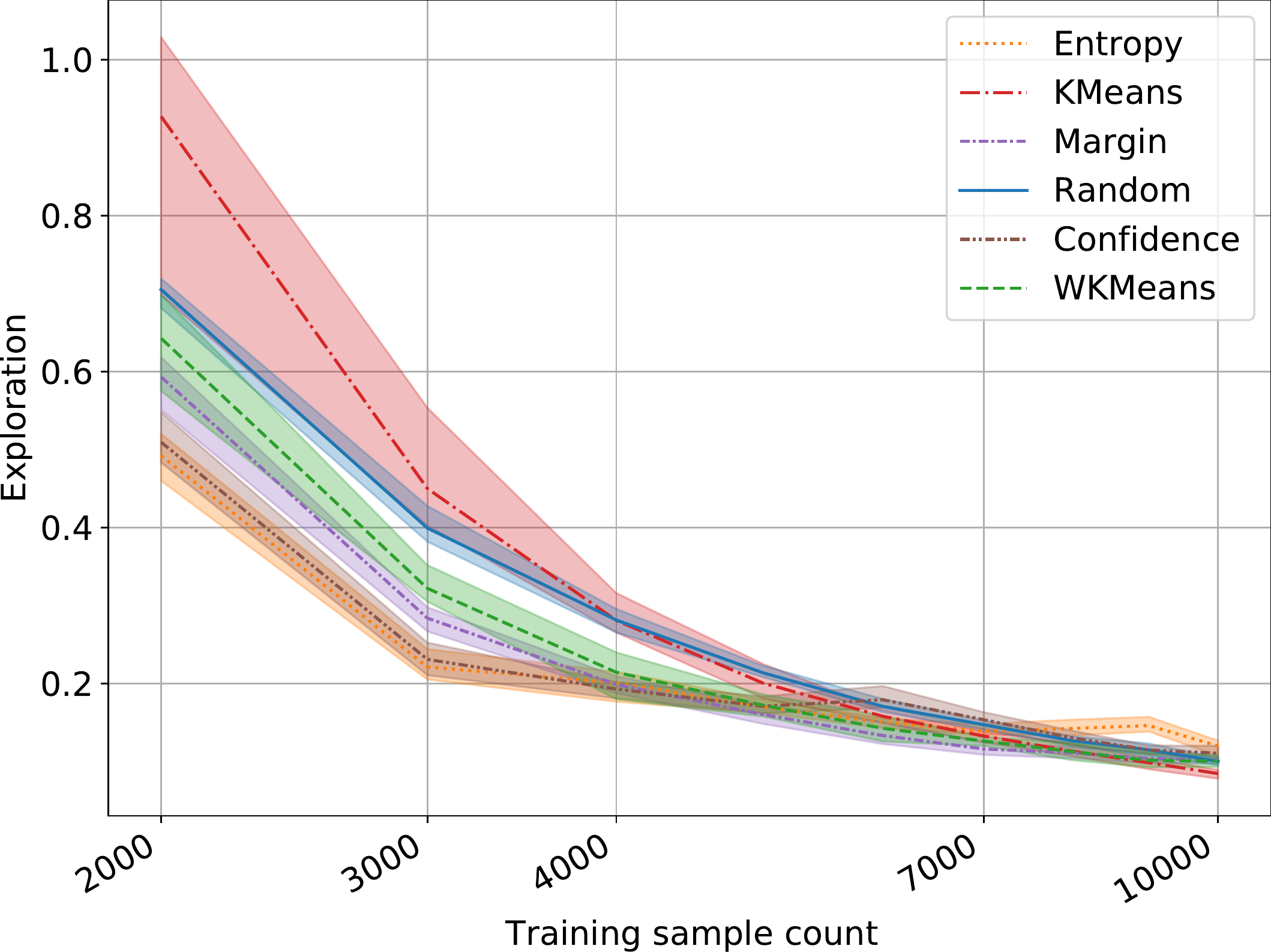}{CIFAR-10}
\exploration{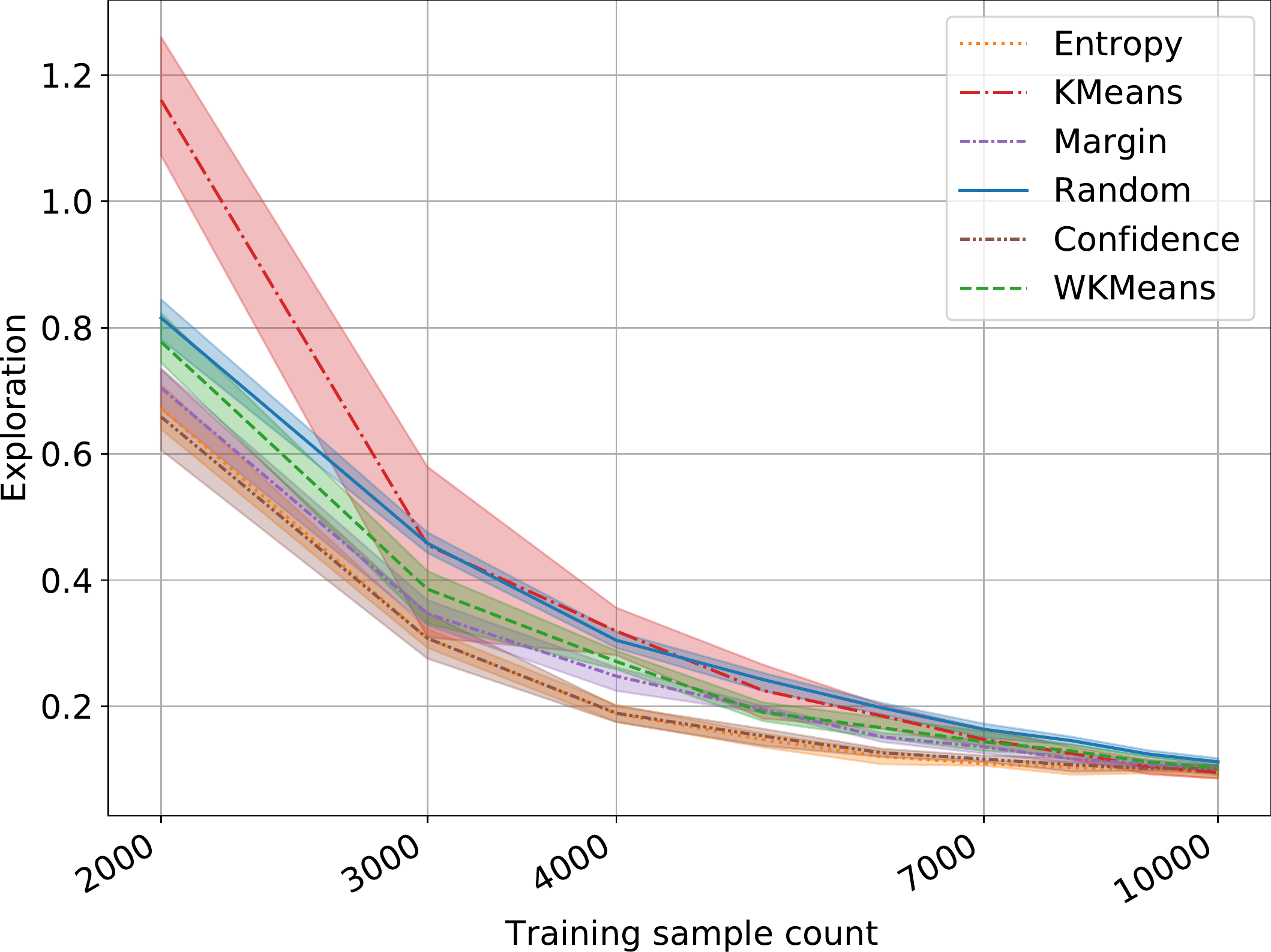}{CIFAR-100}
\exploration{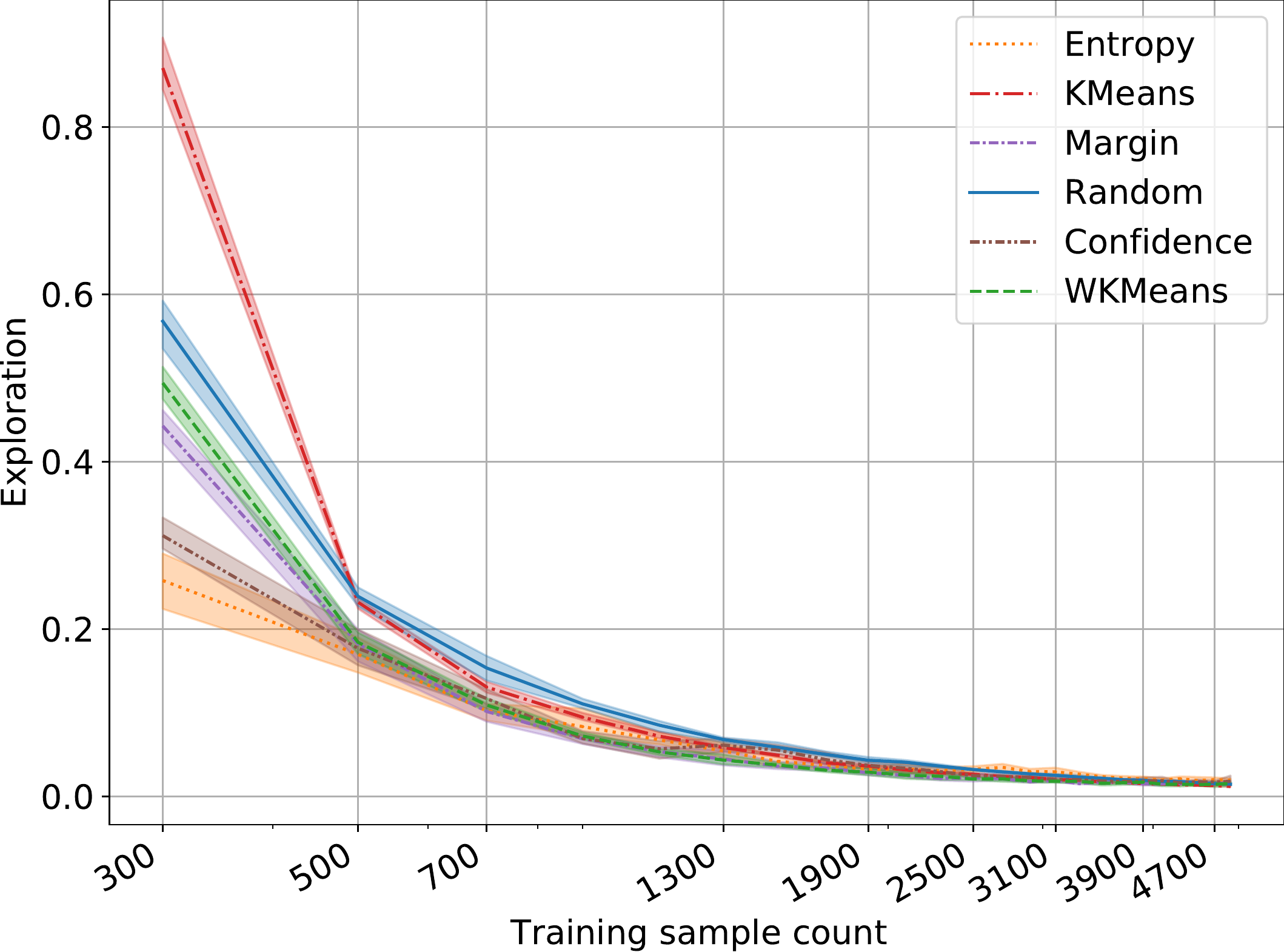}{MNIST}
\exploration{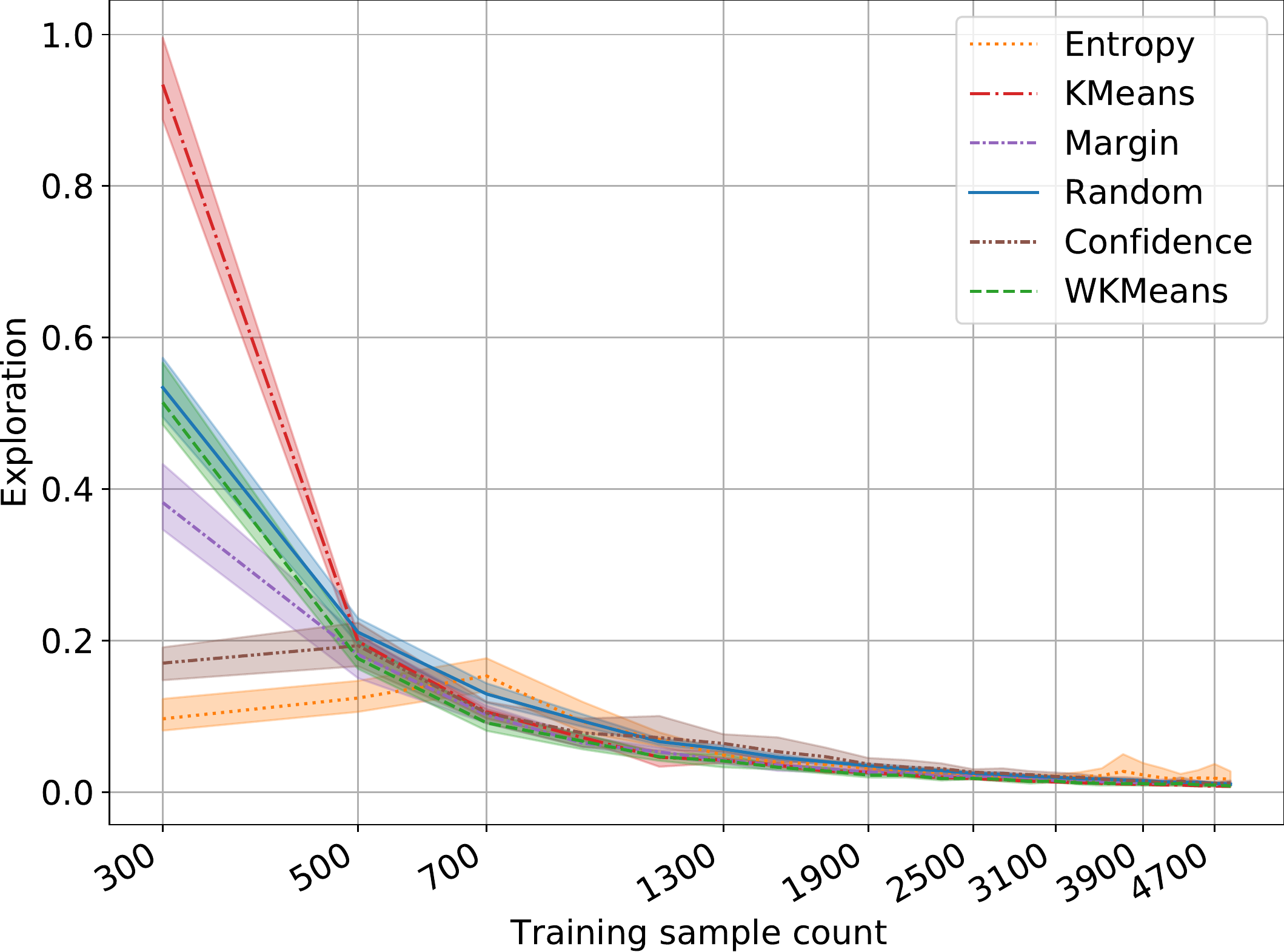}{Fashion}
\exploration{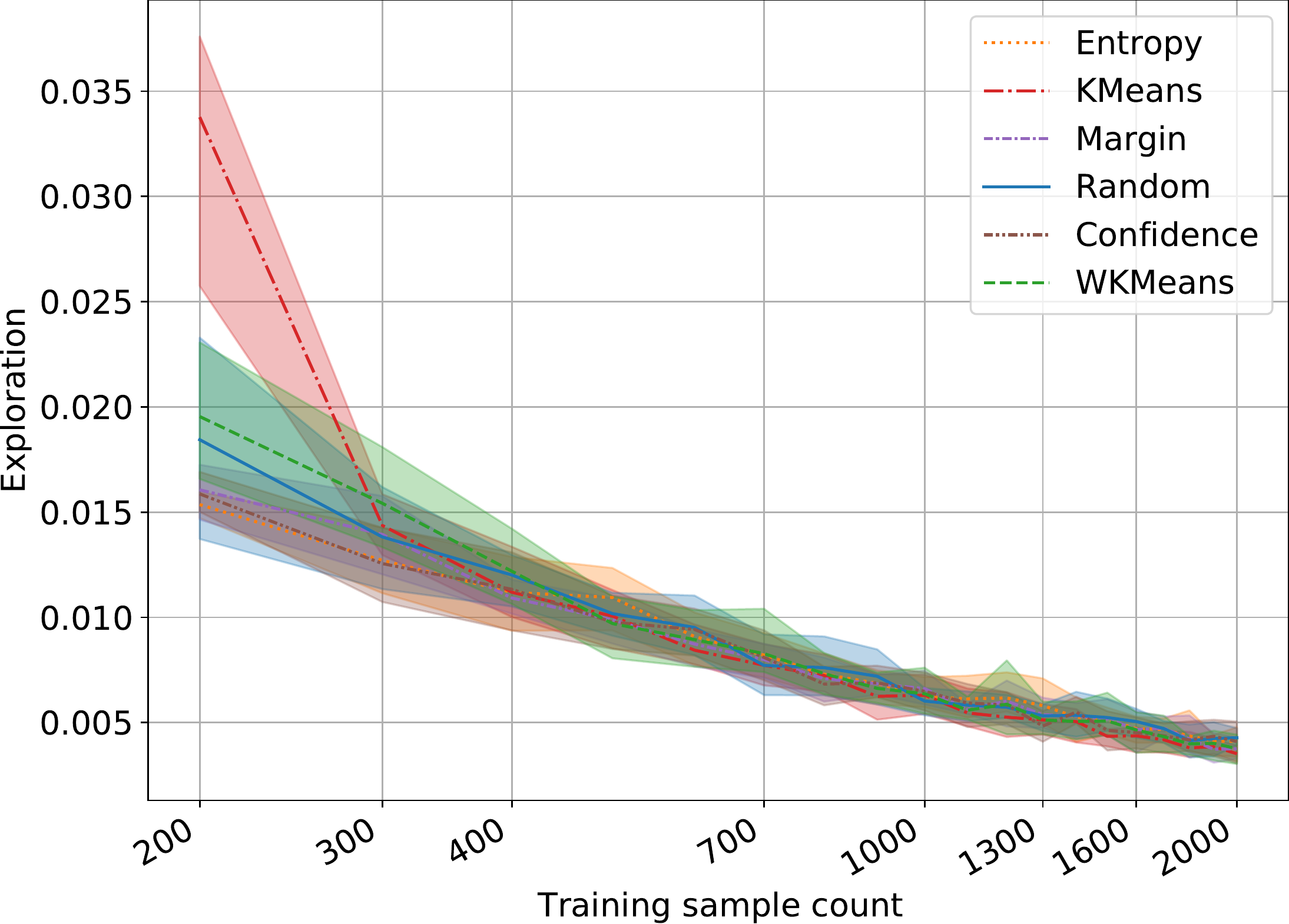}{20 NG}
\exploration{ldpa}{LDPA}
\exploration{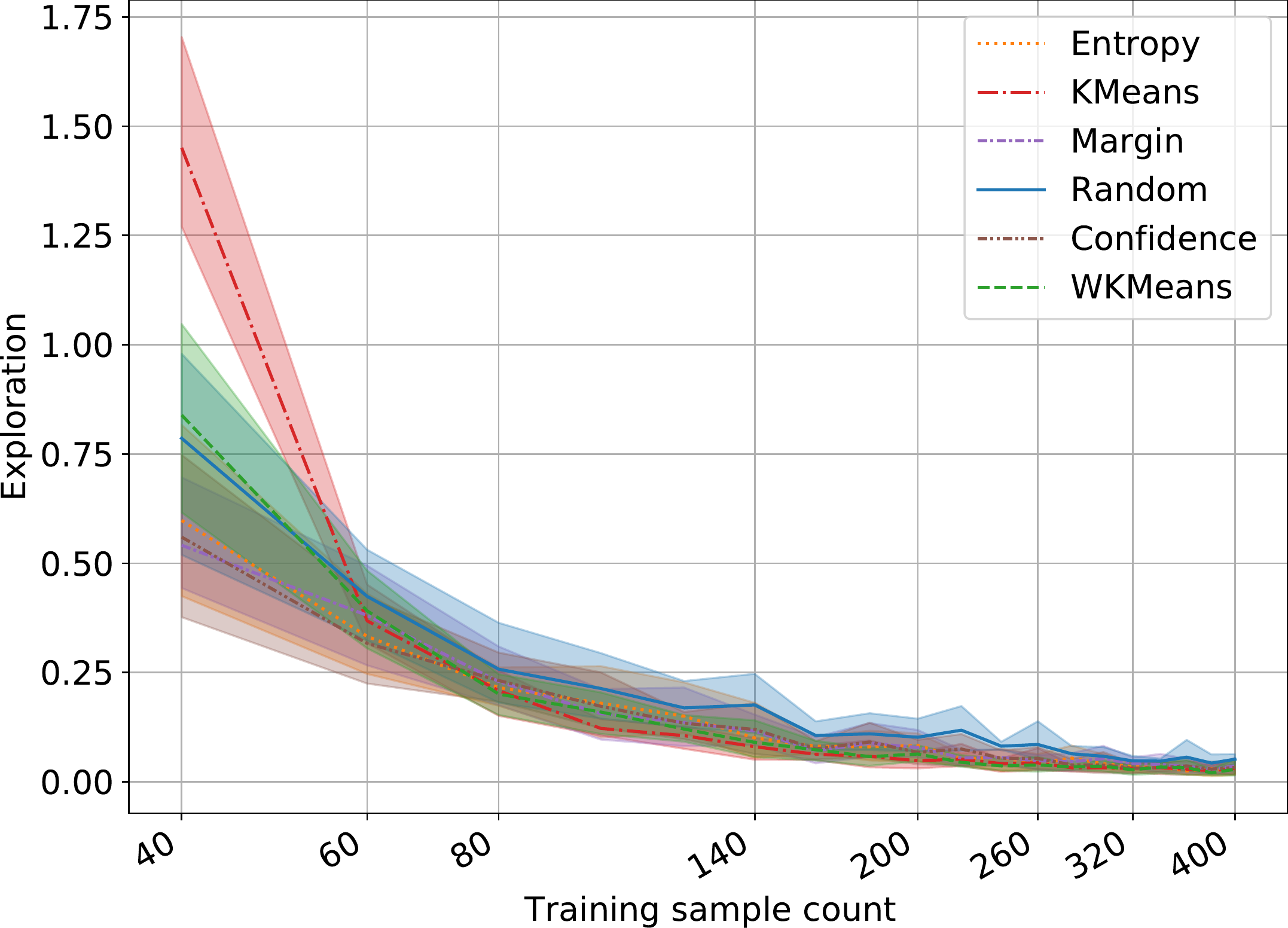}{NOMAO}
\exploration{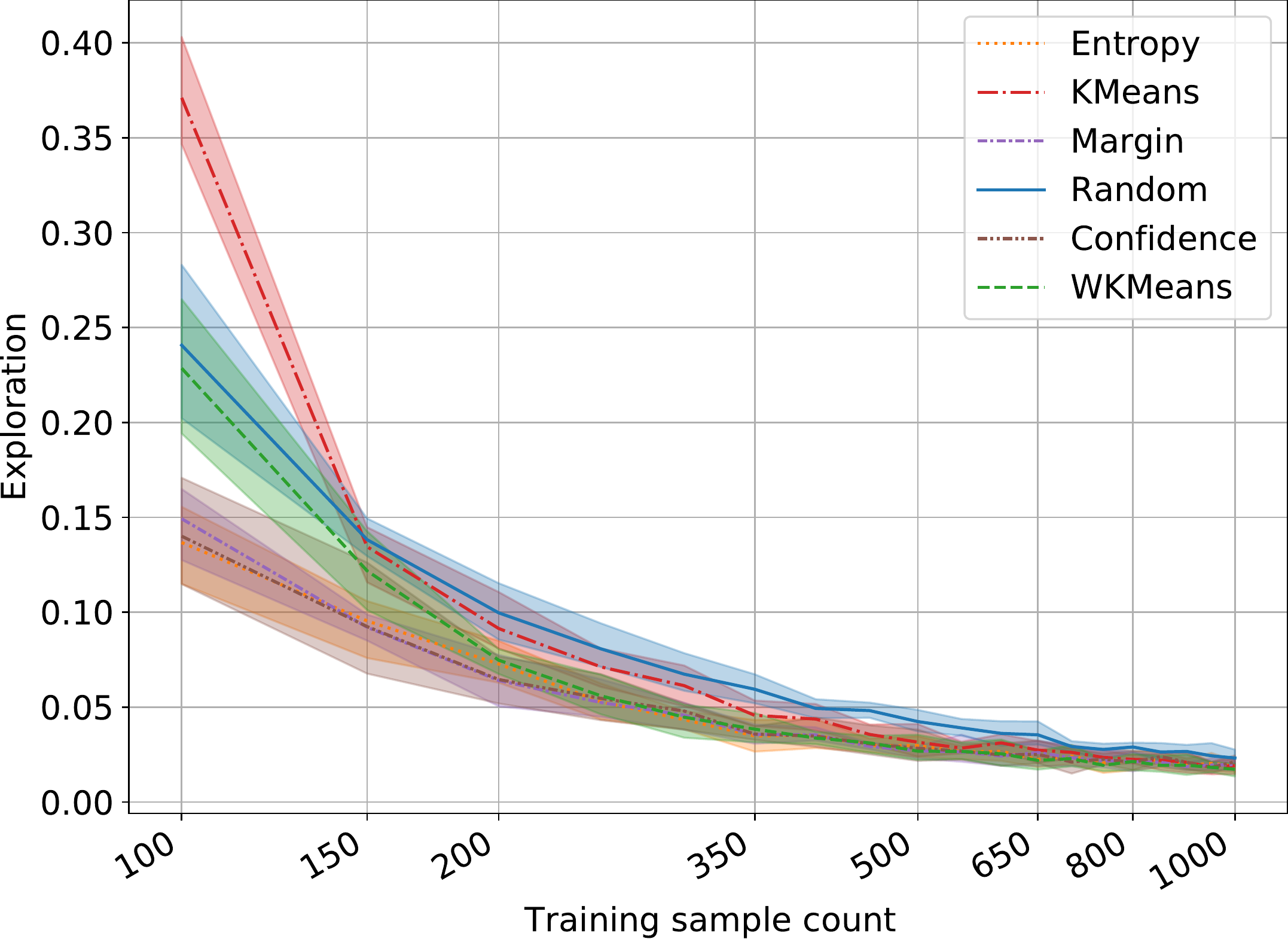}{Phishing}
\exploration{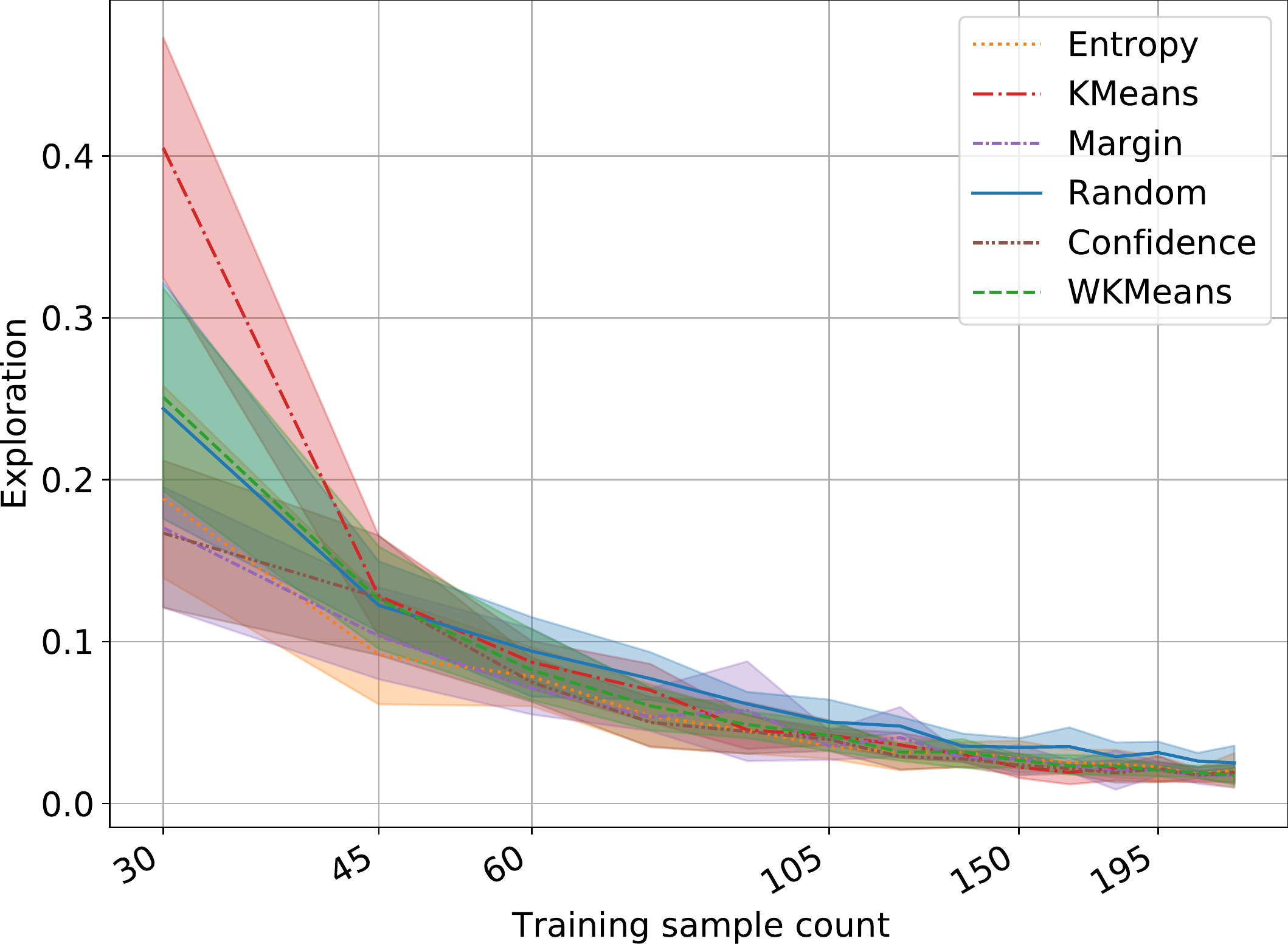}{Robot}

\newcommand{\batch}[2]{
    \begin{figure*}
    \includegraphics[width=.48\linewidth]{figures/#1_batch_difficulty.pdf}\hfill
    \includegraphics[width=.48\linewidth]{figures/#1_batch_agreement.pdf}
    \caption{Reverse batch accuracy (left) and batch classifier agreement (right) for \textbf{#2}}
    \label{app:#1_batch}
    \end{figure*}
}

\batch{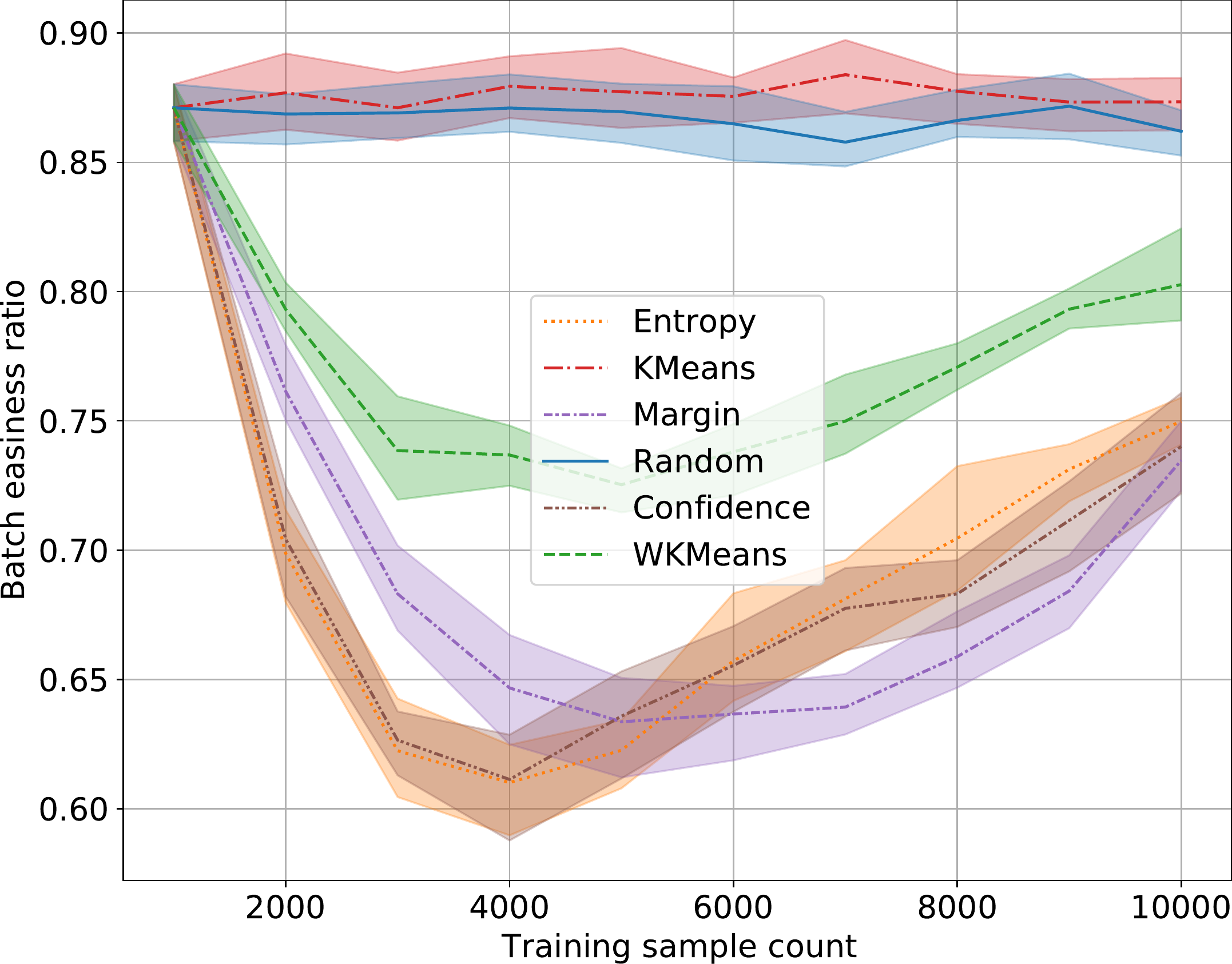}{CIFAR-10}
\batch{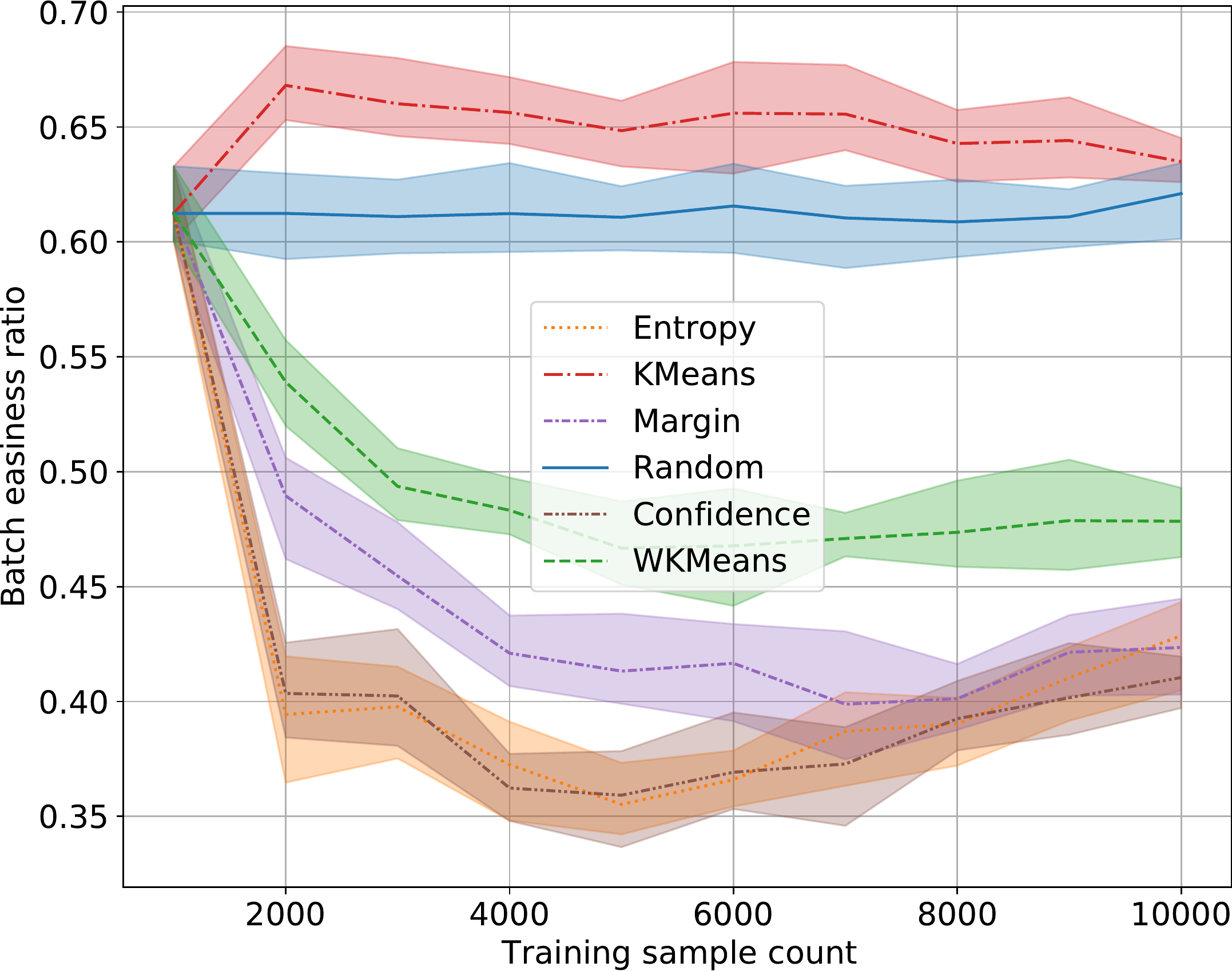}{CIFAR-100}
\batch{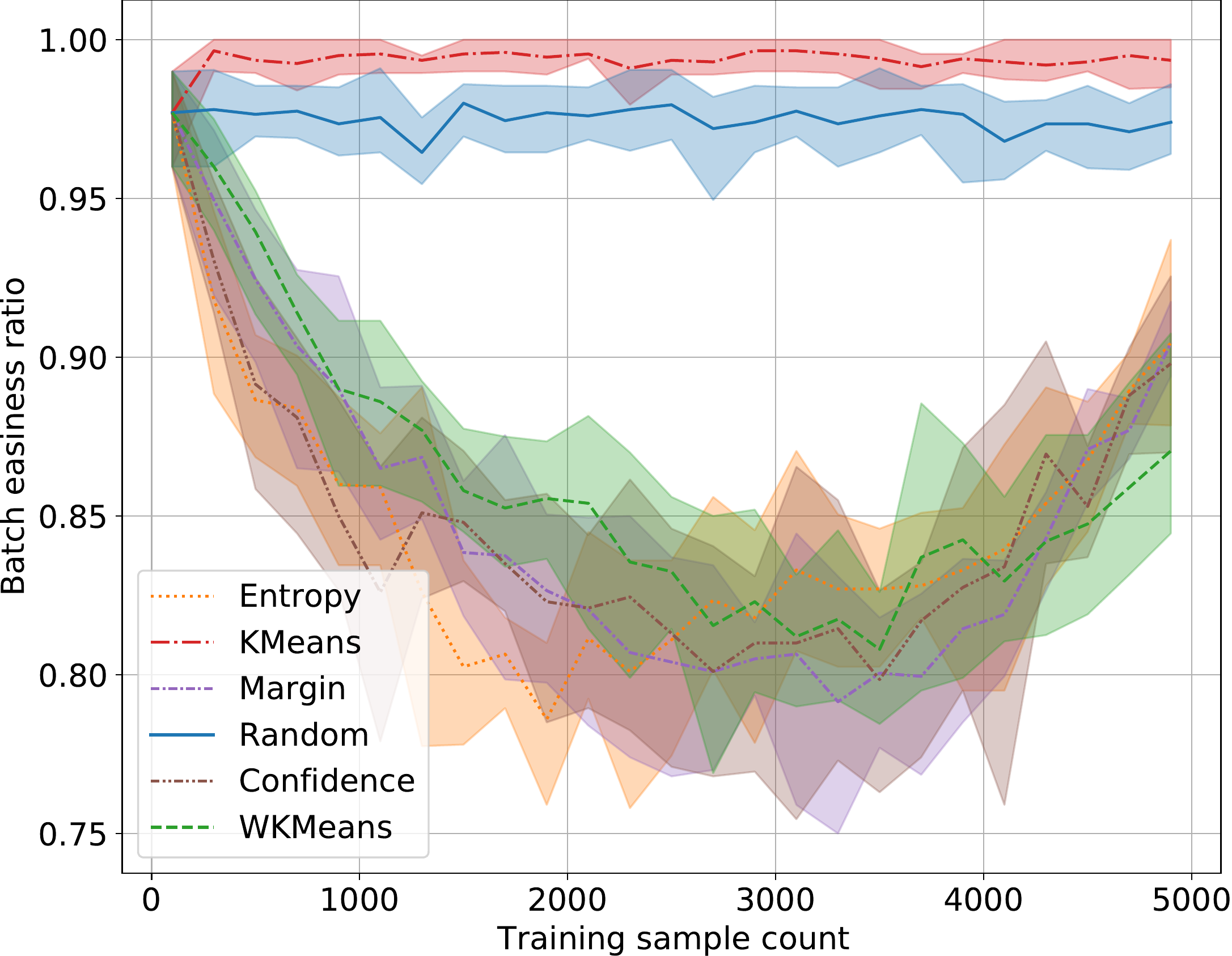}{MNIST}
\batch{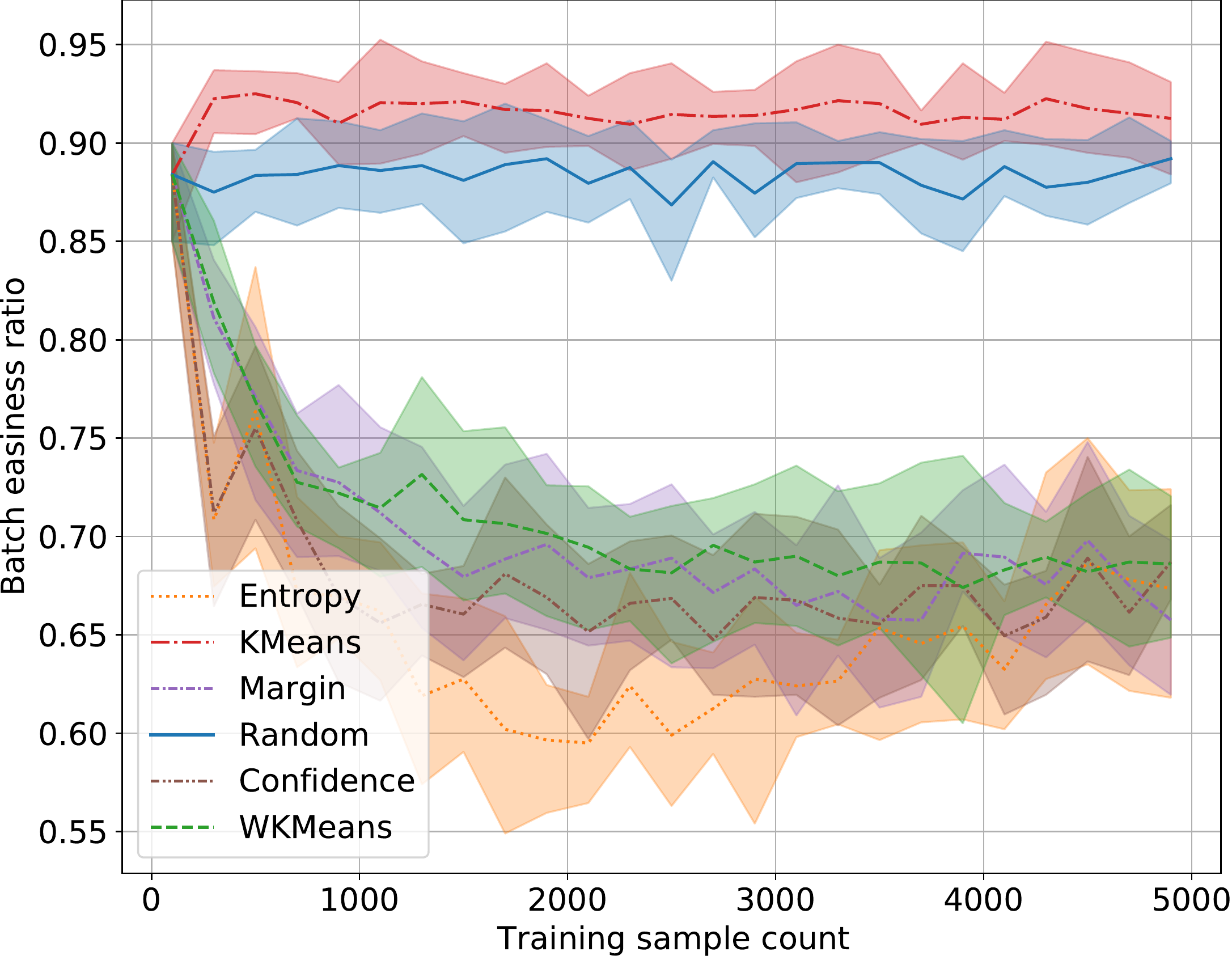}{Fashion}
\batch{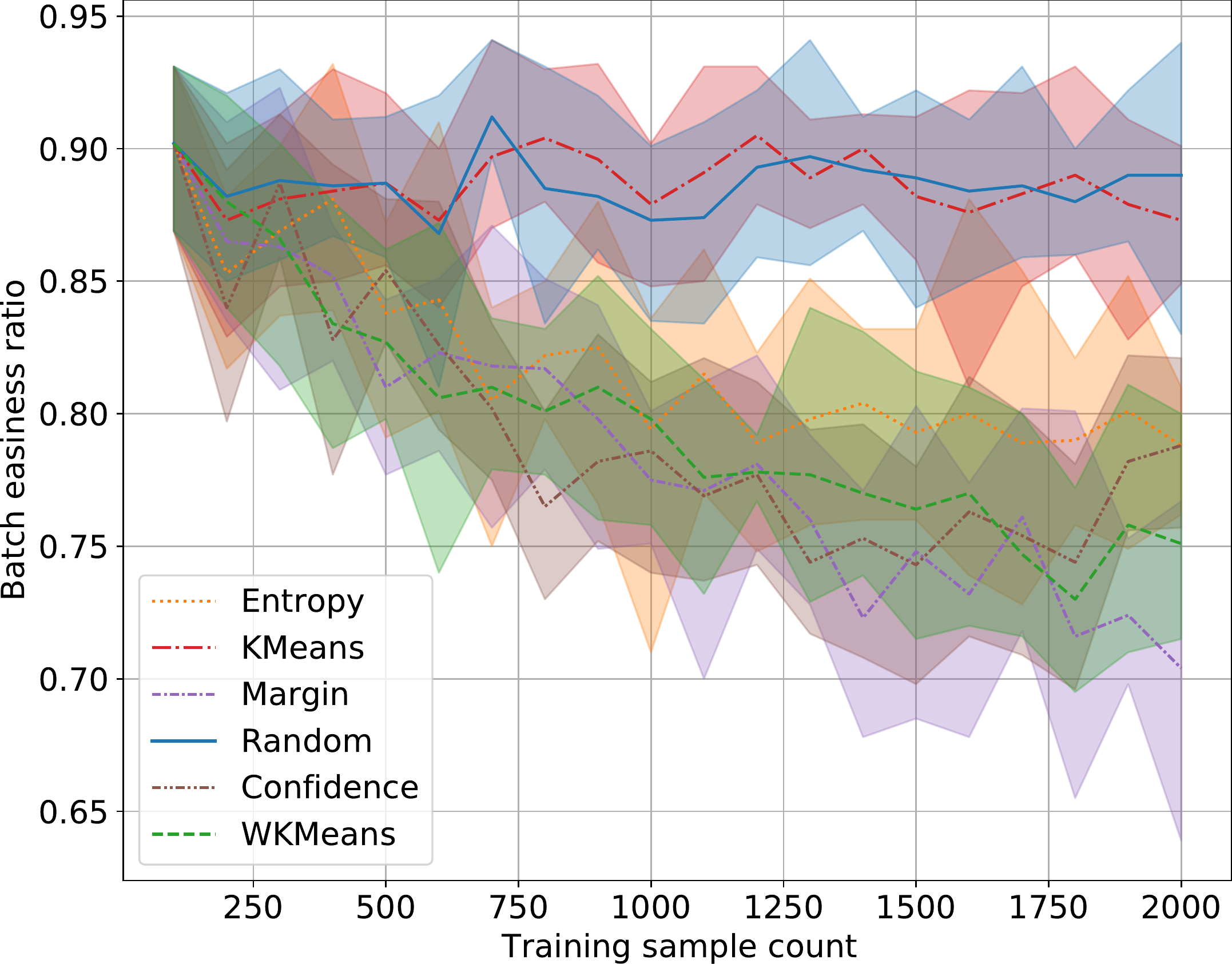}{20 NG}
\batch{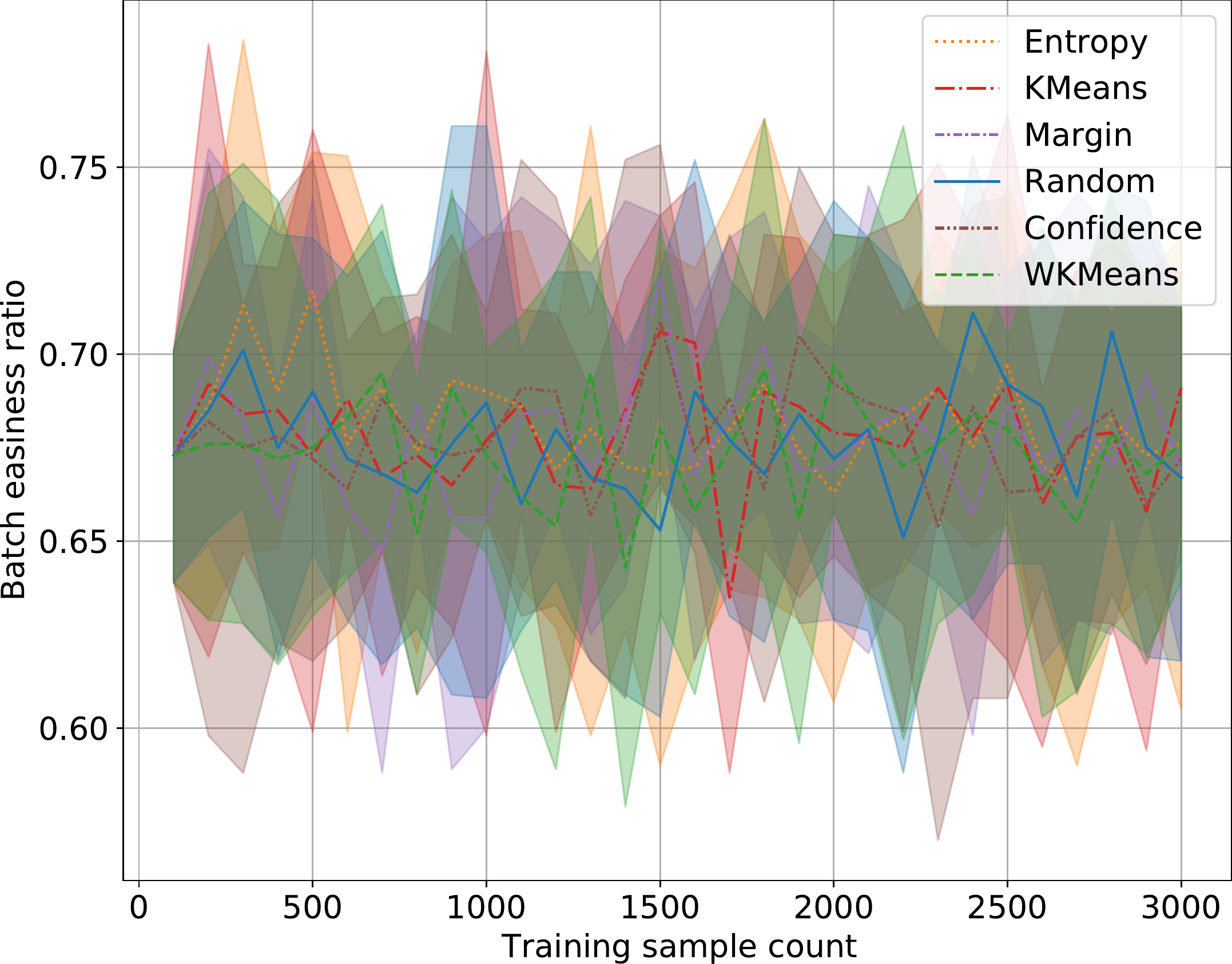}{LDPA}
\batch{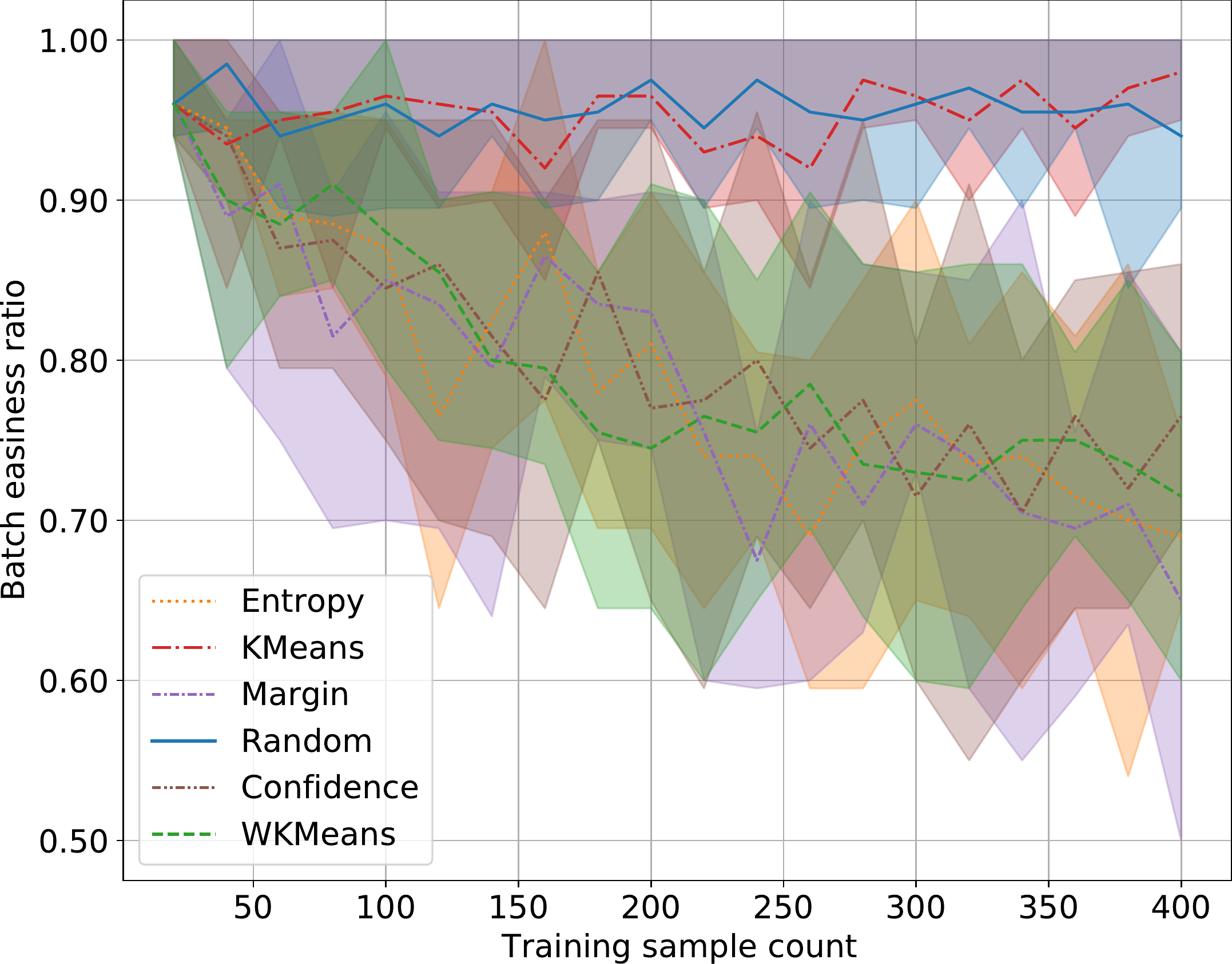}{NOMAO}
\batch{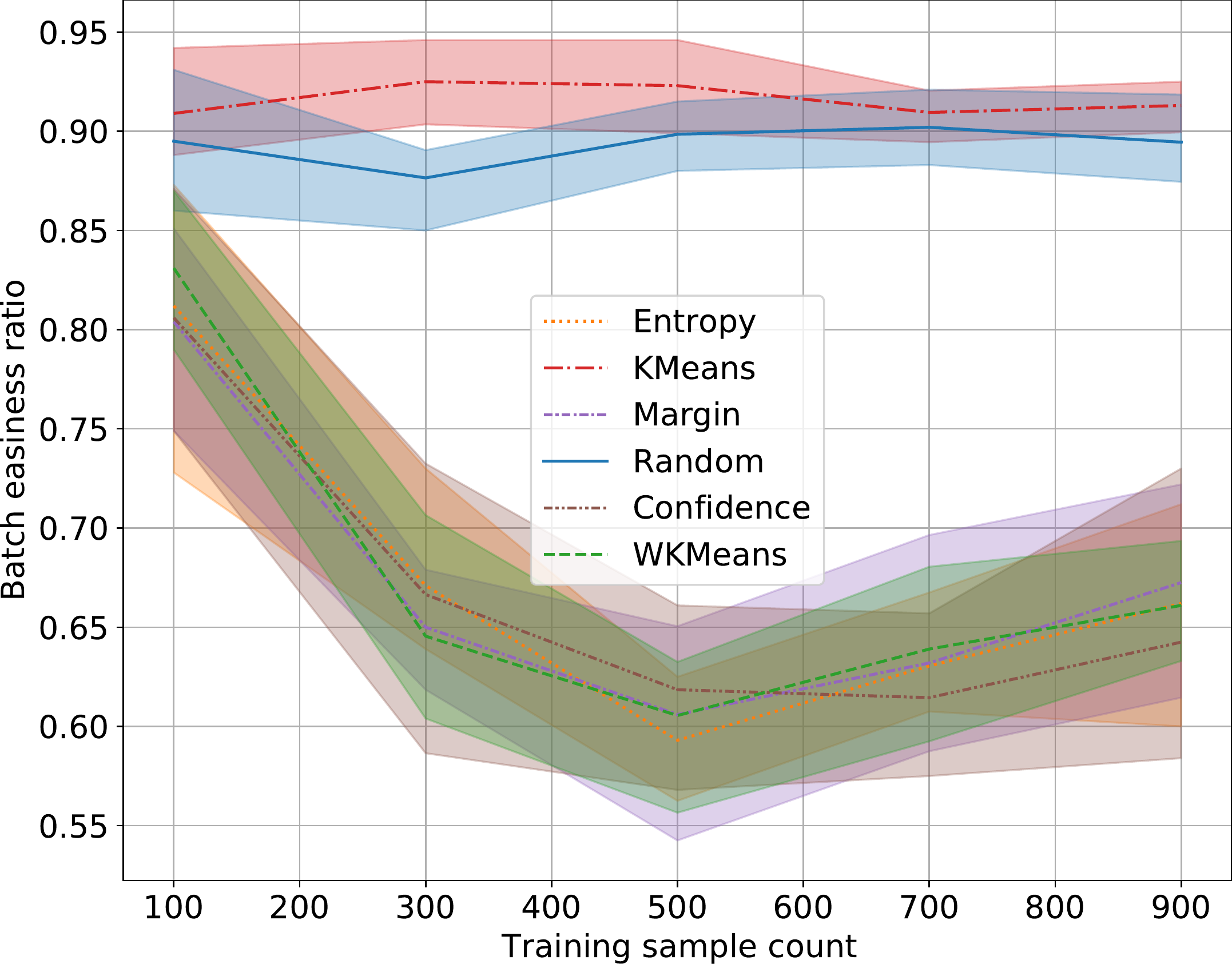}{Phishing}
\batch{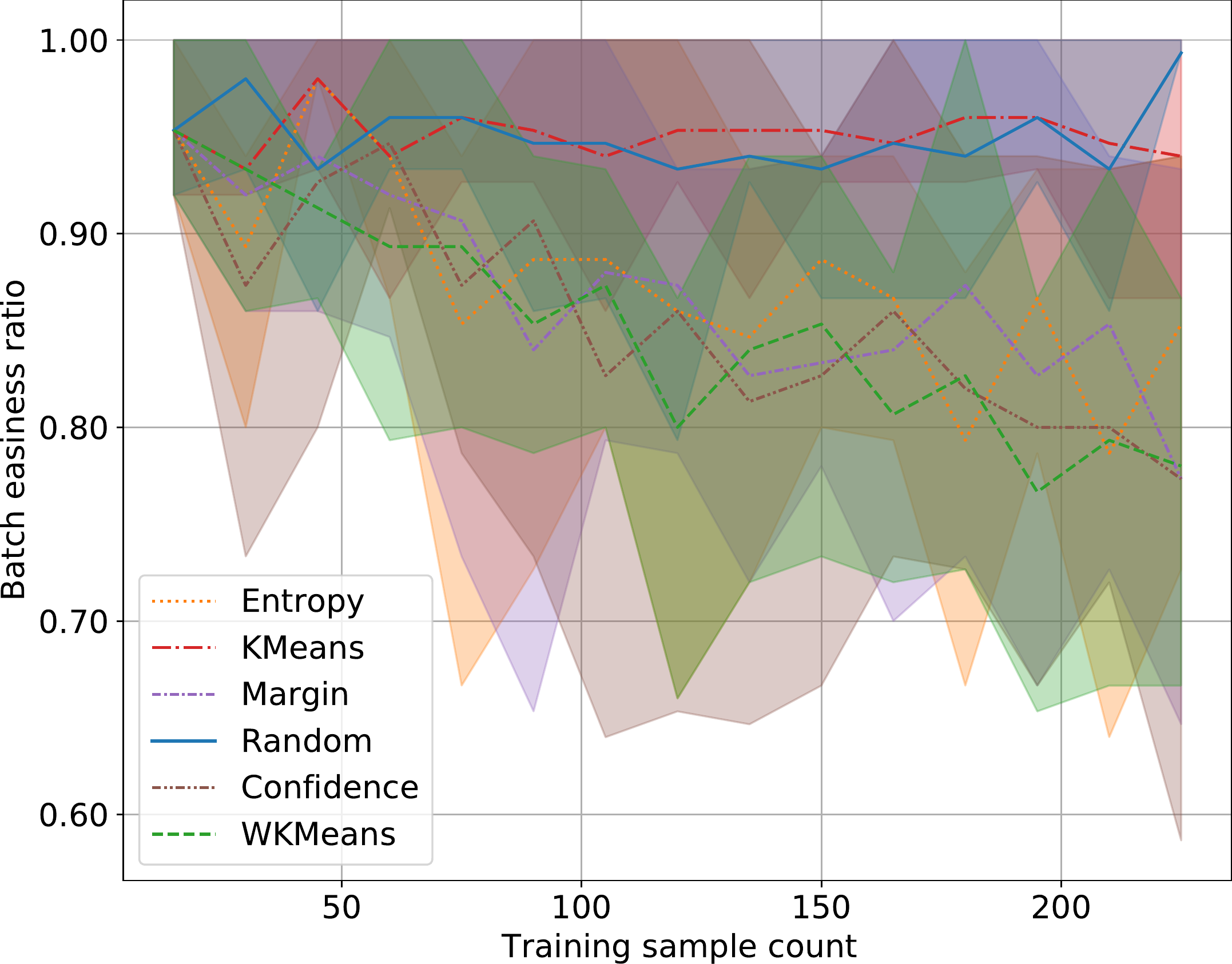}{Robot}

\end{document}